%% file: main.tex
\documentclass[lettersize,journal]{IEEEtran}
\IEEEoverridecommandlockouts                            
\usepackage{breqn}

\usepackage{color}
\usepackage{colortbl}
\usepackage{graphicx} 
\usepackage[american]{babel}
\usepackage{subfigure}		
\usepackage{balance}
\usepackage{cite}
\usepackage{xcolor}
\usepackage{hyperref}
\usepackage{amsmath}
\usepackage{bbding}
\usepackage{bm}
\usepackage{fontawesome}
\usepackage{nicematrix}
\usepackage{array}
\usepackage{dblfloatfix}   
\usepackage{stackengine}

\usepackage{amssymb}

\usepackage{algorithm}
\usepackage{algpseudocode}
\usepackage{multirow}

\title{\LARGE \bf
CLIPSwarm: Generating Drone Shows from Text Prompts with Vision-Language Models 
}

\author{\centering Pablo Pueyo, Eduardo Montijano, Ana C. Murillo and Mac Schwager
\thanks{This work was supported by a DGA scholarship; DGA project T45\_23R, by MCIN/AEI/ERDF/European Union NextGenerationEU/PRTR project PID2021-125514NB-I00; NSF grants CNS-1330008 and IIS-1646921; ONR grants N00014-18-1-2830, and N62909-19-1-2027.}
\thanks{P. Pueyo, E. Montijano and A. C. Murillo are associated with the Instituto de Investigaci\'on en Ingenier\'ia de Arag\'on, Universidad de Zaragoza, Spain 
\texttt{\small \{ppueyor, emonti, acm\}@unizar.es}}
\thanks{M. Schwager is associated with Dept. of Aeronautics and Astronautics, Stanford University, USA
\texttt{\small \{schwager\}@stanford.edu}}
}

\begin{document}

\maketitle
\thispagestyle{empty}
\pagestyle{empty}
\begin{abstract}
This paper introduces CLIPSwarm, a new algorithm designed to automate the modeling of swarm drone formations based on natural language. The algorithm begins by enriching a provided word, to compose a text prompt that serves as input to an iterative approach to find the formation that best matches the provided word. The algorithm iteratively refines formations of robots to align with the textual description, employing different steps for ``exploration" and ``exploitation." 
Our framework is currently evaluated on simple formation targets,  limited to contour shapes. 
A formation is visually represented through alpha-shape contours and the most representative color is automatically found for the input word. 
To measure the similarity between the description and the visual representation of the formation, we use CLIP\cite{clip}, encoding text and images into vectors and assessing their similarity. Subsequently, the algorithm rearranges the formation to visually represent the word more effectively, within the given constraints of available drones. Control actions are then assigned to the drones, ensuring robotic behavior and collision-free movement. 
Experimental results demonstrate the system's efficacy in accurately modeling robot formations from natural language descriptions. The algorithm's versatility is showcased through the execution of drone shows in photorealistic simulation with varying shapes. We refer the reader to the supplementary video for a visual reference of the results.
\end{abstract}


\section{Introduction}
\label{sec_intro}

\input{01_Intro}

\section{Related work}
\label{sec_related}
\input{02_Related_work}

\section{Solution}
\label{sec_problem_formulation}
\input{03_Solution.tex}

\section{Experimental Validation}
\label{sec_experiments}
\input{04_Experiments}

\section{Limitations}
\label{sec_limitations}
\input{05_Limitations}

\section{Conclusions}
\label{sec_conclusions}
\input{06_Conclusions}

\balance
\bibliographystyle{IEEEtran}
\bibliography{references}

\end{document}

%% file: 01_Intro.tex
\textit{Foundation models}, including large language models, image and video generation models, and vision-language models, have revolutionized the technological landscape due to their ability to generate, manipulate, and translate complex natural language and image data. These models are pre-trained on vast amounts of diverse data and can perform a wide array of language-related tasks, including text generation \cite{chatGPT}, translation, and more. Their adaptive nature and applicability show their growing impact, redefining how we interact with technology. This impact can be observed in several key areas such as content creation, customer support, 
software development tasks, and scientific research. In this last realm, foundation models find application in robotics. Existing solutions propose different techniques to control actuators using natural language or giving commands to robots to perform different actions  (e.g. move to the room on your left) ~\cite{huang2023voxposer}. Nevertheless, to the best of our knowledge, there are no prior works in the literature that use foundation models to control a swarm of robots. In this work, we propose to use a vision-language model, CLIP, to drive a swarm of robots to a target formation that matches a text description.

On the other hand, \textit{Artistic robotics} has emerged as a promising field in recent years for both the general audience and the robotics community. The main focus of this trend is using robots to express or design art in various manners, such as the aim to transform robots into painters ~\cite{scalera2019non}, sculptures~\cite{ma2021stylized}, dancers, or even cinematographers~\cite{cinempc}. 
One of the latest trends in robotic arts involves the use of a team of robots or drones that collaboratively create artistic shapes, which is the main focus of this work.  Such ``drone displays," where drones function as pixels and coordinate to create visually appealing shapes in the sky, have become common features of major public sporting and celebration events (e.g. the Olympics Opening Ceremonies, civic New Year's celebrations) with several companies focused on producing these artistic shows. Existing solutions in the literature need the interaction of a person to manually design the shape that the robots should form~\cite{nar2022optimal}.

\begin{figure}[!t]
\centering
\textbf{\small{Text Prompt: \textit{``Leaf"}}}\par\medskip
\begin{tabular}{c}  
\includegraphics[width=0.35\columnwidth]{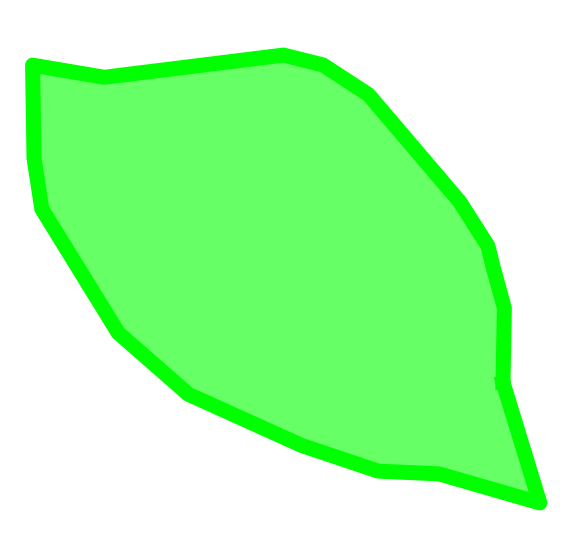}

    \includegraphics[width=0.62\columnwidth]{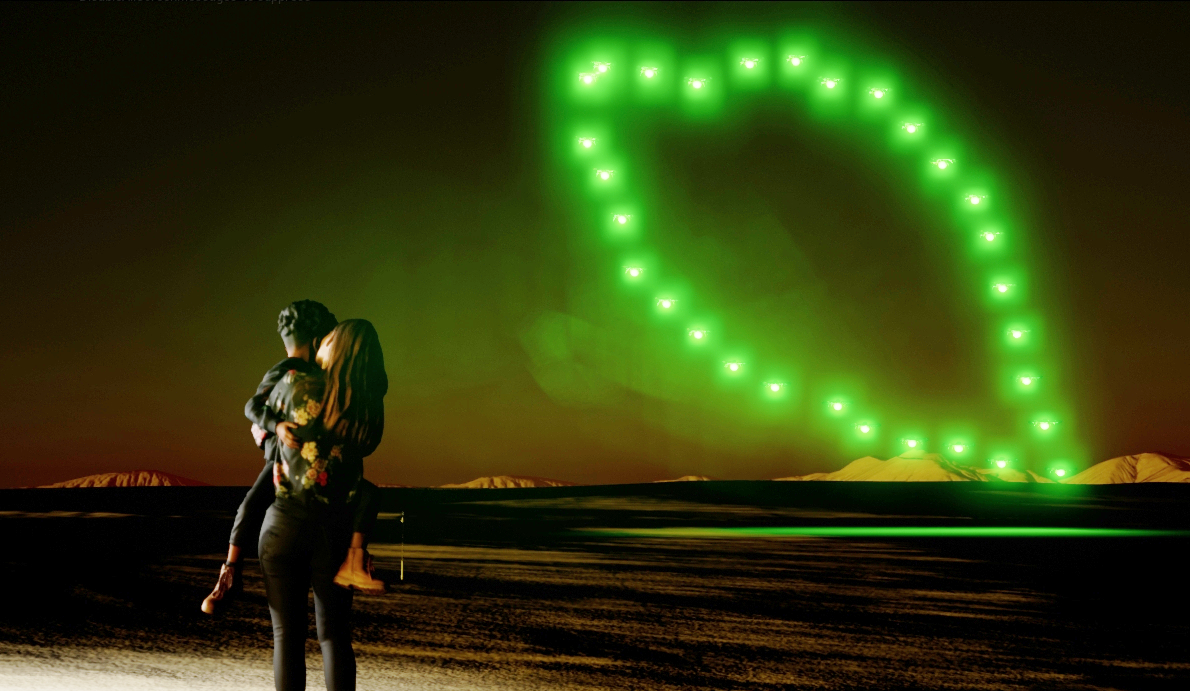}
\end{tabular}
\caption{\footnotesize{\textbf{Drone formation automatically crafted to match a given text}. CLIPSwarm takes a single word describing a shape as input and determines automatically the color and positions of a robotic swarm formation that best fits the given text. The example illustrates the shape created by a formation of 30 robots. The drones move to positions that collectively form a shape corresponding to the word \textit{``Leaf''}. } Left: graphical representation of the shape formed by the robot formation. Right: formation of drones as part of a show in a photorealistic simulation representing the input word.}
\label{fig:main}
\end{figure}

Complementing the existing literature on the aforementioned topics, the primary objective of this project is to utilize an existing \textit{foundation model} for shaping a swarm of robot formations with \textit{artistic intent}. CLIPSwarm paves the way and is the first step to creating robot formations autonomously. As shown in Fig.\ref{fig:main}, the user introduces a description in natural language of the desired shape of the formation, e.g.\textit{ ``leaf"}, and CLIPSwarm automatically decides the position and color of the robots of the formation to form a shape and a drone formation that corresponds with that description, so the users do not need to create the patterns beforehand. 
To do so, we use CLIP ~\cite{clip}, a multimodal foundation model that is trained to encode texts and images, finding similarities between them, and calculating what is referred to as CLIP Similarity. The system takes as input a text prompt describing a desired shape, and the output comprises the positions of the robots to best represent that shape, along with the color that is best suited to the prompt. 

To achieve this goal, our proposed method begins with an engineered prompt. The algorithm selects the most representative color for the input text and elaborates an enhanced prompt to provide additional information to CLIP. 
Then, we execute an iterative algorithm to choose the best formation describing the introduced text, improving the CLIP Similarity across iterations. As the last step, the system identifies the most representative robot positions that form the desired shape and their corresponding colors.  As output, our method generates 
the robot locations and color, which are then fed to a lower-level distributed formation control system to drive the robots to their goal positions while avoiding collisions. This user-friendly approach facilitates the process for users, eliminating the need to pre-create patterns, and offers a dynamic and efficient means of translating artistic descriptions into robotic formations.

Our method is validated with extensive experiments. We first analyze how the proposed algorithm increases the CLIP similarity obtained through iterations, and many qualitative examples showcase how the shape achieved by the formations matches the input text. 
We also run a simulated drone show to represent the process of creating several figures in photorealistic simulation, employing a navigation algorithm to avoid collisions between robots and giving realistic robotic dynamics to the drones. The video is included as part of the supplementary material. The presented experiments demonstrate CLIPSwarm's applicability to real robotic systems, and its ability to create swarm formations of robots that correspond to given descriptions.

%% file: 02_Related_work.tex
Foundation models, such as large-scale language models, have seen widespread adoption in various fields, including robotics. Researchers have been exploring ways to integrate natural language understanding and generation capabilities into robotic systems \cite{firoozi2023foundation, zeng2023large, zhang2023large}. Some applications include the utilization of multimodal foundation models to control manipulators \cite{huang2023voxposer}, direct actions \cite{cui2022can}, or design a plan of actions to a robot performing a particular task \cite{huang2022language}. Certain works integrate foundation models with reinforcement or imitation learning techniques, achieving human-like behavior in robots \cite{kim2024guide, wake2023gpt}. One of the latest solutions employs ChatGPT~\cite{chatGPT} to directly generate Python code capable of controlling heterogeneous robots to perform various actions as commanded in natural language \cite{vemprala2023chatgpt}.  While existing solutions represent a significant leap in the application of foundation models for robotics, CLIPSwarm stands out as the first solution utilizing a foundation model to control a cooperative swarm of robots with artistic purposes without the need for retraining or fine-tuning any model.

Artistic robotics is an emerging research area in robotics that is gaining attention from both the general audience and the robotics community. Various works explore the transformation of robots into painters, with techniques ranging from replicating existing styles of real painters \cite{scalera2019non, beltramello2020artistic} to preprocessing input images for simplified robot input \cite{karimov2021image}, or even autonomously creating graffiti \cite{chen2022gtgraffiti}. Some works focus on robots sculpting without human interaction \cite{ma2021stylized}. Artistic expressions extend beyond sculpture or painting; for example, some solutions involve robots acting as professional dancers \cite{peng2015robotic} or serving as autonomous cinematographers. In this latter application, robots autonomously record cinematic scenes, satisfying artistic or technical details \cite{bonatti2020autonomous, cinempc, pueyo2023cinetransfer}.

In the realm of multirobot artistic robotic formations, some works employ optimal control to move 2D formations of robots forming a set of given patterns, \cite{alonso2011multi, alonso2012image}. The same authors incorporated an interface to draw the desired pattern in \cite{hauri2013multi}. More recent approaches address how to perform multidrone formations shows, \cite{nar2022optimal,  waibel2017drone, kim2016realization, sun2020path}. All these works receive the desired pattern as input. Recent works generate 3D shapes using diffusion models \cite{luo2021diffusion} or GANs \cite{hui2020progressive} but the behavior of the particles, which do not  explicitly consider robotic behaviours and constraints, 
can make it difficult to adapt to a robotic extent.

In contrast to the aforementioned approaches, CLIPSwarm allows users to give simple instructions to the system in natural language, without the need to create patterns beforehand. This is achieved thanks to the use of Foundation Models, eliminating the need for extensive datasets or time-consuming retraining. The platform automatically determines the best positions and color for the team of robots to represent the formation described by the text input. It then provides an approach to performing a drone show, considering the dynamics and potential collisions inherent in a robotic swarm system.

%% file: 03_Solution.tex
\begin{figure*}[!h]
\centering
\begin{tabular}{c}  
\includegraphics[width=2\columnwidth]{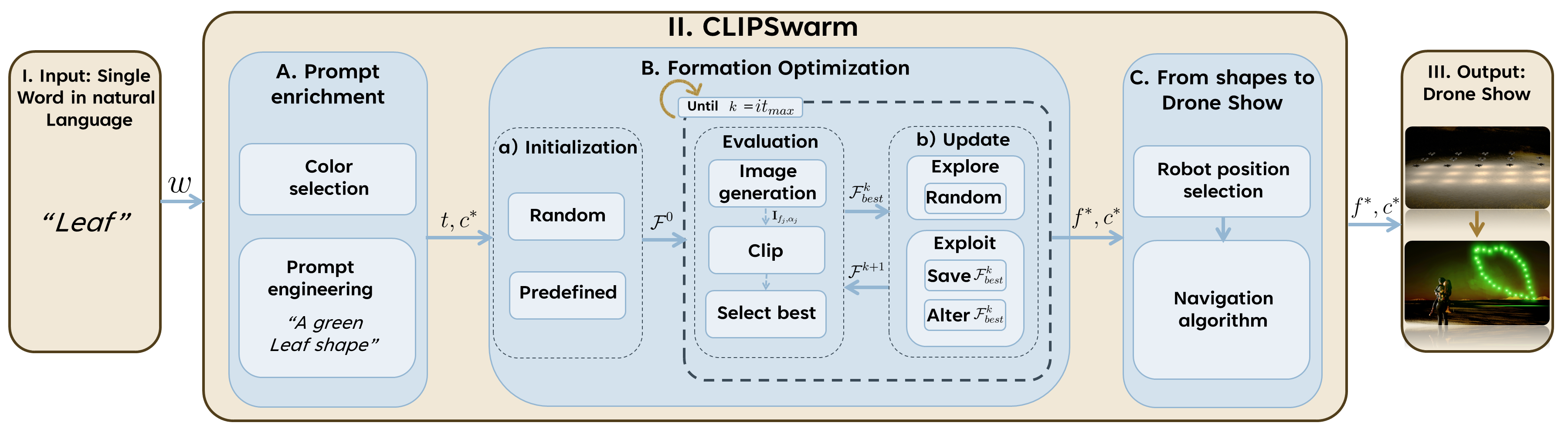}

\end{tabular}
\caption{\footnotesize{\textbf{CLIPSwarm algorithm diagram}.} A schematic summary of the platform, its modules, and their interactions. \textit{I. Input of the system}, which is a word describing the desired formation. \textit{II. CLIPSwarm algorithm}, including the three modules of the system. \textit{(A) Prompt enrichment}, involving color selection and prompt engineering to enrich the input word and form a text. \textit{(B) Formation Optimization}, incorporating the steps select the formation that best describes the input text. \textit{(a) Initialization}. A set of formations (consisting of robot position) are randomly sampled from a uniform distribution. Some predefined shapes are added as part of a 'warm start'. \textit{Evaluation}. The formations are converted to images. Then, CLIP extracts the similarity between the images and the input text, and the formations with the best similarities are selected. \textit{(b) Update}. New formations are iteratively created employing an "exploration-exploitation" strategy, improving the CLIP similarity across iterations. \textit{(C) From shapes to drone show}. The positions of the obtained formation are optimized through robot position selection and a navigation algorithm. \textit{III. Output of the system} is drone positions to perform a drone show by moving and selecting the color of the drones representing a shape described by the input word.}
\label{fig:diagram}
\end{figure*}

Our solution is organized into three distinct modules, as illustrated in Fig. \ref{fig:diagram}. The first module enriches the input word by incorporating additional details to create a more accurate text prompt. The second module employs an iterative algorithm to enhance the \textit{CLIP Similarity} between the prompt and the images associated with a set of formations.
This metric is formulated using CLIP~\cite{clip}, a foundational model trained on an extensive dataset of text-image pairs. 
We denote by $\text{CS}:t\times\mathbf{I}\rightarrow [0,1]$ the CLIP similarity function, that returns a positive score measuring the similarity between an input text $t$ and an image $\mathbf{I}$.
This function achieves higher values when the text and image pair describe the same concept or idea.
Finally, the third module adapts the output formation and is in charge of assigning goal positions to the drones and move them avoiding obstacles.
The three modules are detailed in the following subsections.

\subsection{Prompt enrichment}
The input of the system is a single word, denoted by $w$. This first module finds the most representative color $c$ and composes an enriched text, $t,$ for more detailed information on the desired shape.
After conducting various tests and comparing different additional words, we observed that the CLIP Similarity is higher when certain words are added to the input text. Specifically, we perform prompt engineering by enriching the input text to describe both a shape and its corresponding color, such as ``A green leaf shape". 

To accomplish this, we start by generating a set of 10 images, $\mathcal{C}=\{\mathbf{I}_1,\ldots, \mathbf{I}_{10}\}$, with all their pixels of a single basic color\footnote{\footnotesize{red, orange, yellow, green, cyan, blue, purple, magenta, pink, brown}}.
We select the color with the highest CLIP Similarity when compared to the input word $w$, 
\begin{equation}
c^* = \arg  \mathop{\max}\limits_{c\in\{1,\ldots,10\}} \text{CS}(w, \mathbf{I}_c),
\end{equation}
considering it as the most representative for the input word. Subsequently, we add the remaining words to construct a text prompt $t$ in the following manner:  $t =$ ``A $c^*$ $w$ shape". This text $t$ and the color $c^*$ are then inputted into the second module of our solution for further processing.

\subsection{Formation optimization}
\label{sec_algorithm}

Let $M$ be the number of robots available to create the formation. A formation is defined by the positions of all the robots in the image plane, $f=\{\mathbf{p}_1,\ldots,\mathbf{p}_M\}$, where $\mathbf{p}_i\in\mathbb{R}^2$ is the position of the $i$-th robot of the formation.
The goal is to find the formation, $f^*$, that maximizes the CLIP similarity of $t$ and the image associated with the formation.

\subsubsection{Image generation}
\begin{figure}[!h]
\centering
\begin{tabular}{ccc}
   \includegraphics[width=0.25\linewidth]{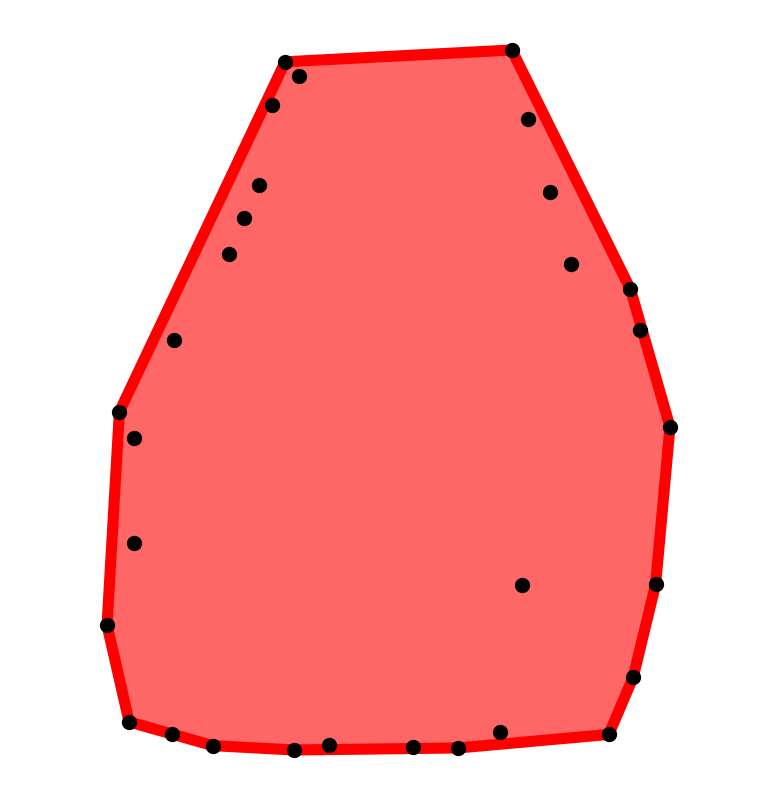}
    & \includegraphics[width=0.25\linewidth]{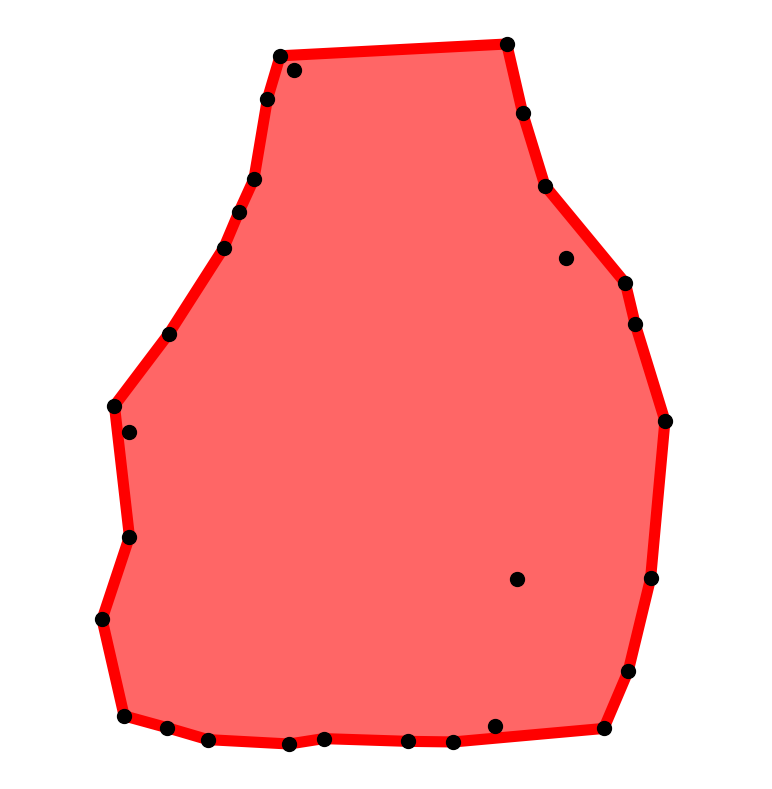}
    & \includegraphics[width=0.25\linewidth]{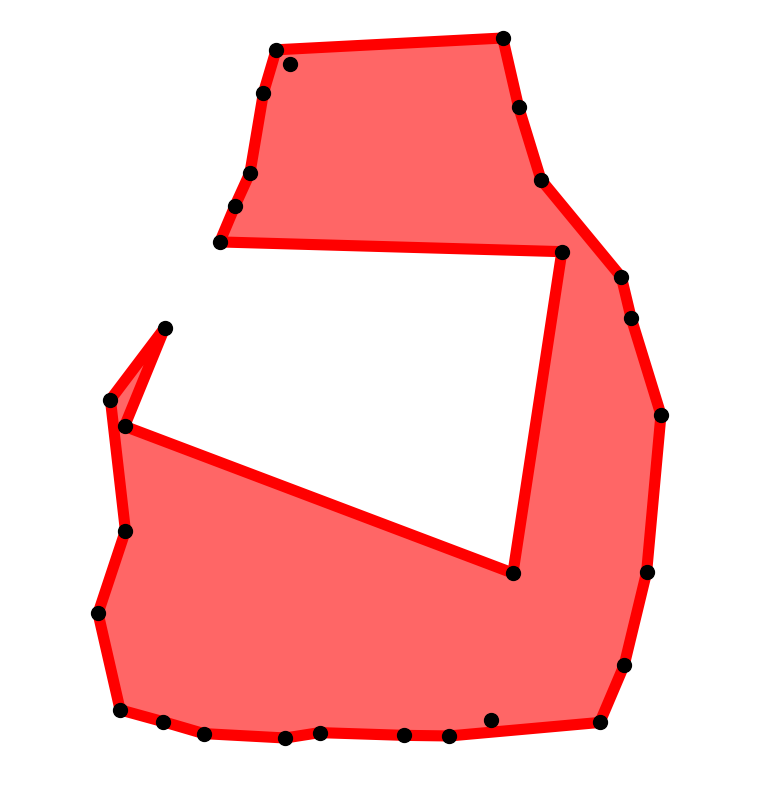}
    \\   
$\alpha=0$ & $\alpha=0.3$ & $\alpha=\alpha_{f}=0.37$ \\ 
\end{tabular}
\caption{\textbf{Influence of Alpha Value on Contour:} Alpha shapes representation of a robot formation with varying alpha values. Different alpha values produce distinct representations of the contour. When the alpha value is zero, the contour forms a convex hull. Larger alpha values result in a more finely detailed and intricate contour. Each formation has a maximum value of alpha ($\alpha_f$) that ensures all points are inside a single polygon while the contour has the maximum concavity.}
\label{fig:alphas}
\end{figure}

Since a robot formation is a sparse collection of points, a fundamental element in our solution is how we generate the images to compute the CLIP similarity.
In this work, images are crafted by drawing a concave contour over the robots positioned on the edge of the formation using the Alpha-Shape algorithm \cite{edelsbrunner2011alpha}. 
The algorithm receives a set of points, i.e., a formation, $f$, and a real number, $\alpha$, as inputs and returns a graph, $G$, where the nodes are the points and the edges depend on $\alpha$.
Particularly, for $\alpha=0$ the algorithm returns a graph with the convex hull of the formation. As $\alpha$ increases, the figure associated with the graph presents a higher concavity, by including more of the formation points in the contour. To obtain a single polygon, the parameter can increase until a limit value, $\alpha_{f},$ depending on $f$, where the contour contains all the points in the formation (Figure~\ref{fig:alphas}).
Bigger values of $\alpha$ than $\alpha_f$ will return a graph with more than one connected component.

Our images are generated by filling the polygon associated with $G$ of the color $c^*$. For simplicity in the notation, we denote by $\mathbf{I}_{f,\alpha}$ the image obtained using this procedure for the formation $f$ and the parameter $\alpha$.
Figure~\ref{fig:formations} in the experiments show different examples of such images.

\subsubsection{Iterative optimization}
Once we are able to compute images from formations, we can use them to compute the best one in terms of the CLIP similarity.
Particularly, we propose an optimization algorithm where we iteratively refine a set of $N$ potential formations and $\alpha$ parameters, based on the CLIP similarity of their corresponding images with respect to $t$.
Let
\begin{equation}
    \mathcal{F}^k = \{ \mathcal{F}^k_1, \ldots, \mathcal{F}^k_N\},
\end{equation}
be the formation set at iteration $k$, where $\mathcal{F}^k_j=\{f^k_j, \alpha^k_j\}, j=1,\ldots,N,$ is a formation and $\alpha$-parameter duple.
Therefore, our process iteratively finds
\begin{equation}
\label{Eq:OptimalForm}
    \{ f^*, \alpha^* \}= \arg\max_{\mathcal{F}_j=\{f_j, \alpha_j \}\in \mathcal{F}^k} CS(t, \mathbf{I}_{f_j,\alpha_j}).
\end{equation}

\begin{figure}[!bh]
\centering
\begin{tabular}{cccccc} 
\includegraphics[width=0.12\linewidth]{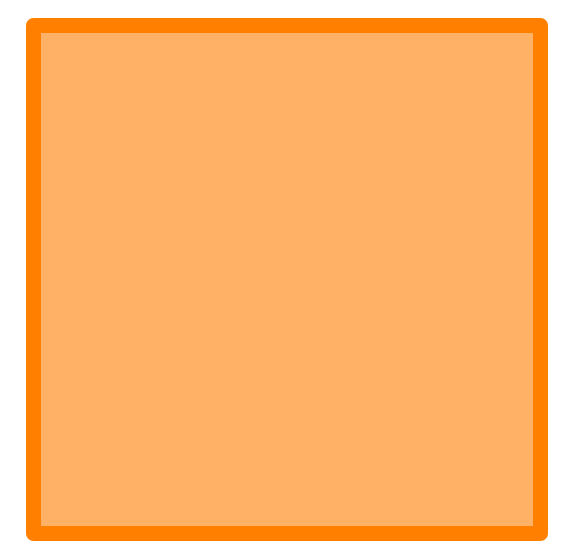}
  & \includegraphics[width=0.12\linewidth]{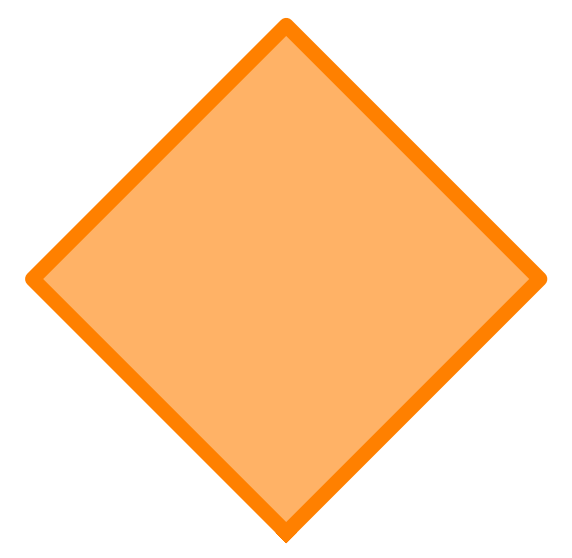}
    & \includegraphics[width=0.12\linewidth]{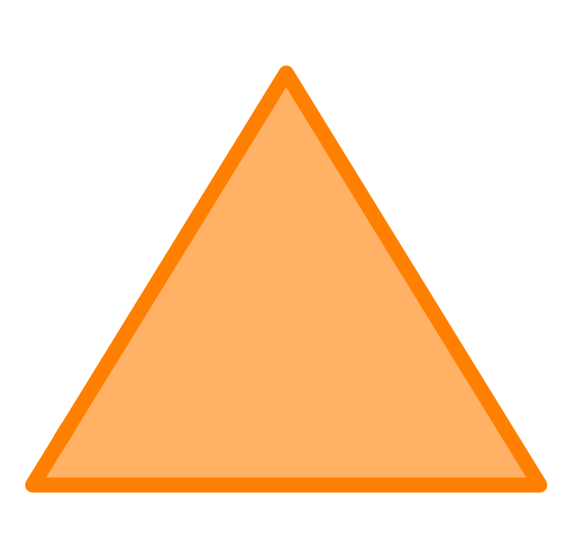}
    & \includegraphics[width=0.12\linewidth]{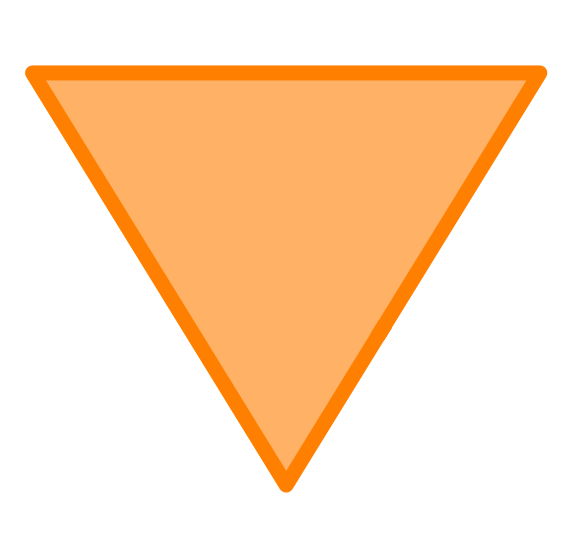}
   & \includegraphics[width=0.12\linewidth]{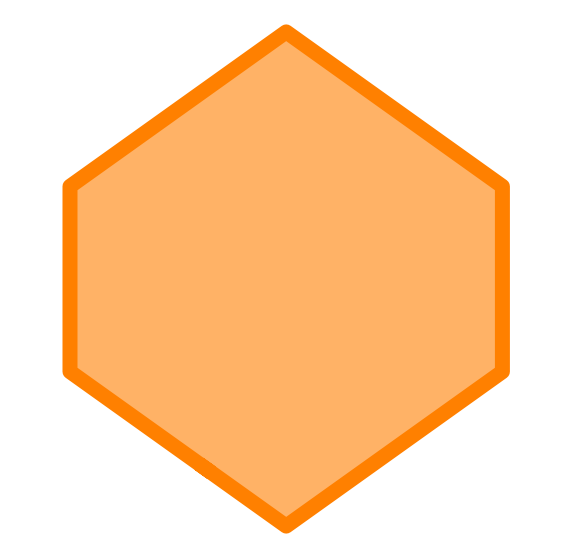}
   & \includegraphics[width=0.12\linewidth]{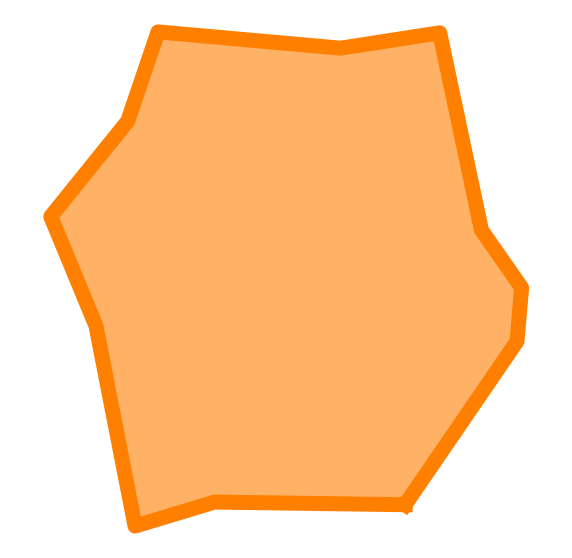}
    \\ 
\end{tabular}
\begin{tabular}{cccccc}\includegraphics[width=0.12\linewidth]{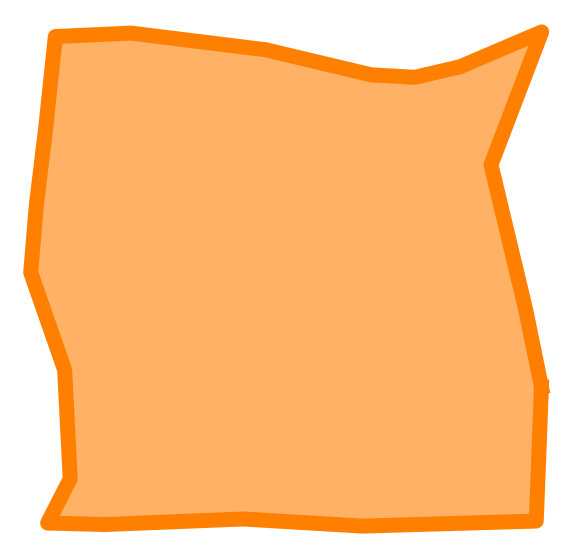}
  &
   \includegraphics[width=0.12\linewidth]{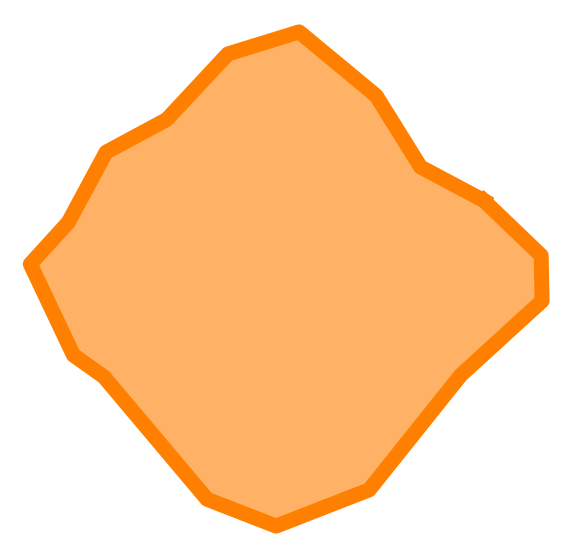}
    & \includegraphics[width=0.12\linewidth]{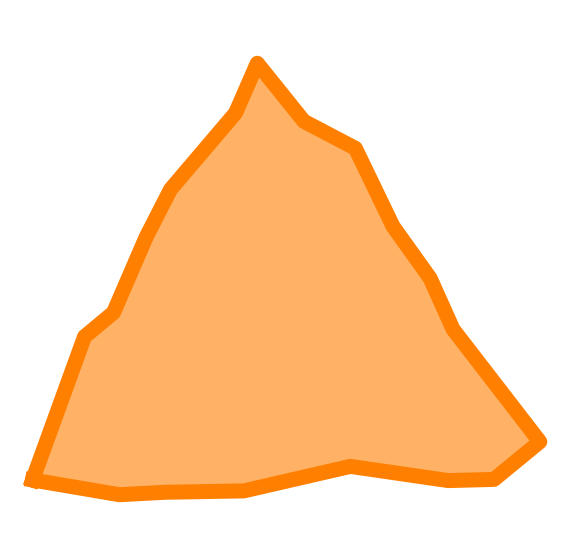}
    & \includegraphics[width=0.12\linewidth]{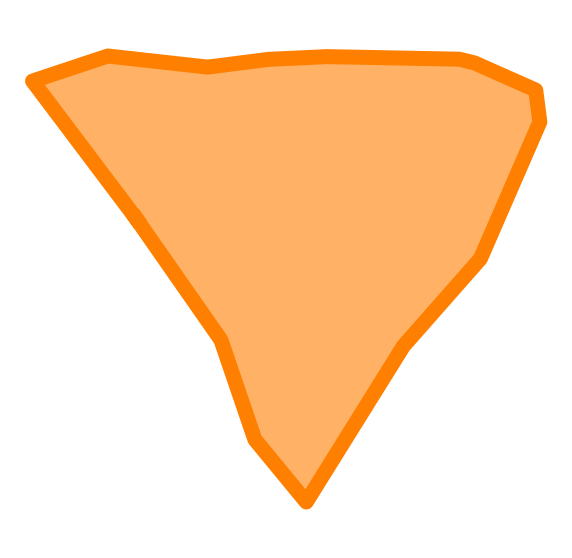}
   & \includegraphics[width=0.12\linewidth]{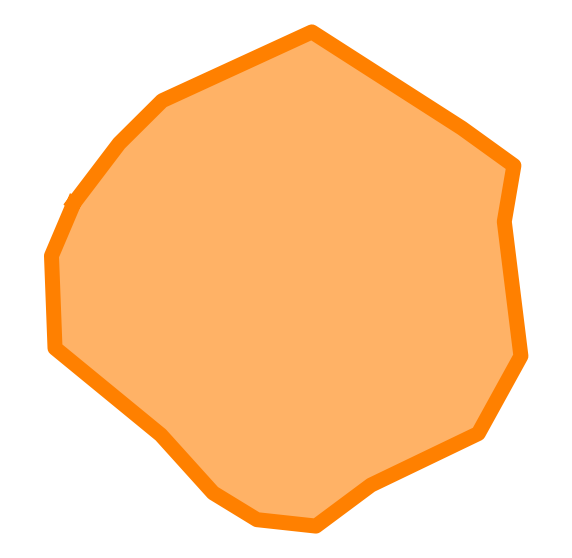}
   & \includegraphics[width=0.12\linewidth]{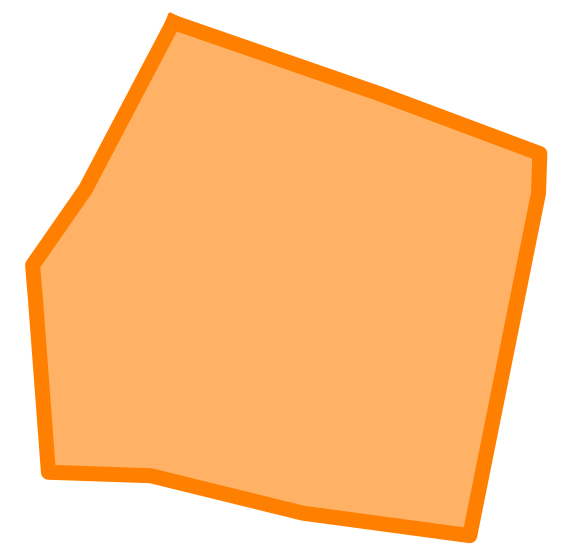}
    \\ 
\end{tabular}
\begin{tabular}{cccccc}\includegraphics[width=0.12\linewidth]{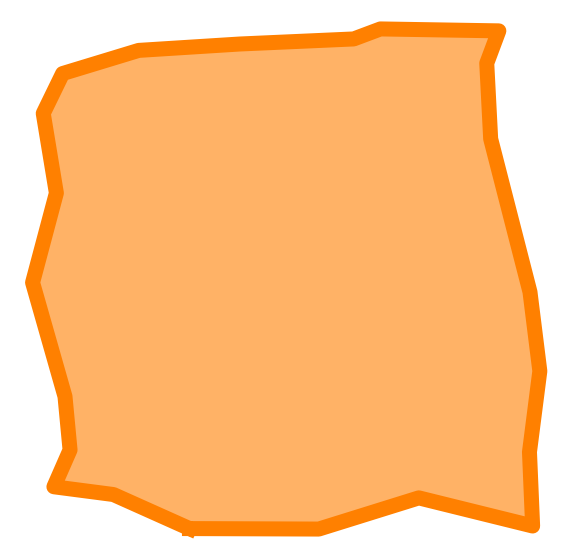}
  &
   \includegraphics[width=0.12\linewidth]{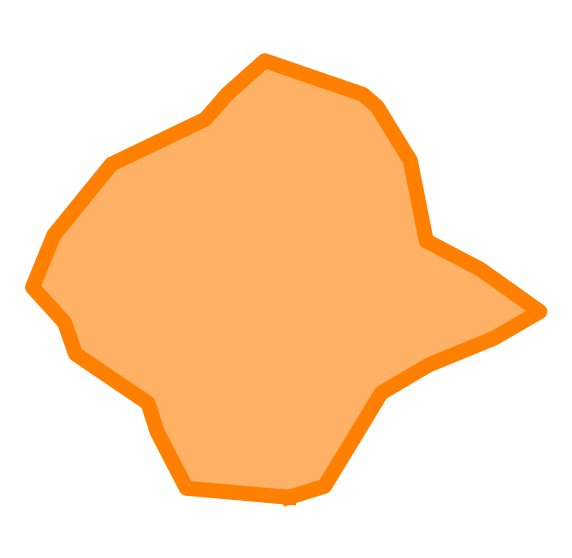}
    & \includegraphics[width=0.12\linewidth]{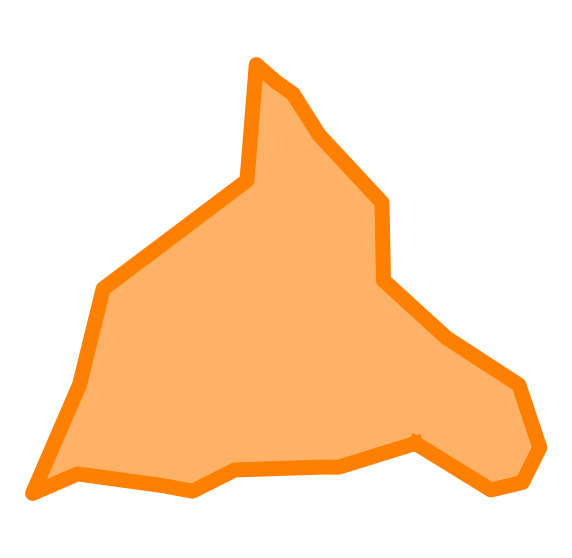}
    & \includegraphics[width=0.12\linewidth]{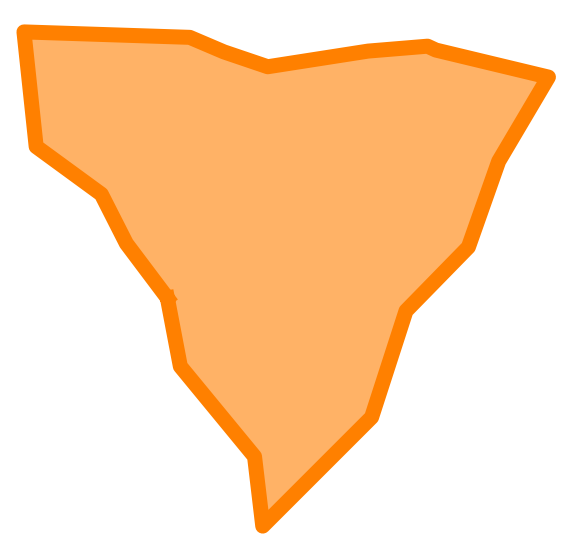}
   & \includegraphics[width=0.12\linewidth]{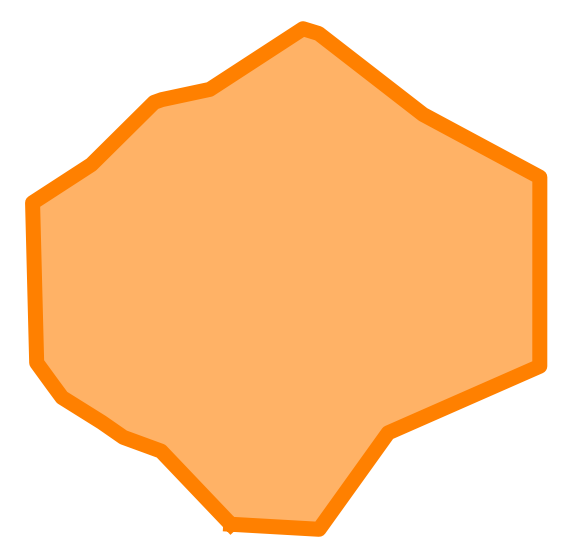}
   & \includegraphics[width=0.12\linewidth]{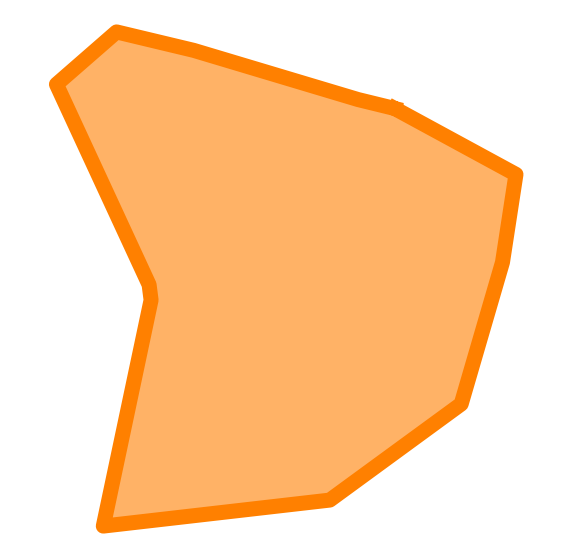}
    \\ 
\end{tabular}
\caption{\textbf{Predefined shapes}. Columns 1-5 display predefined shapes along with random variations of them, which are added to the initialization pool as a 'warm start' during the Initialization stage. Column 6 shows some random samples from the initialization pool for comparison.
}
\label{fig:predefined}
\end{figure}

\paragraph{Initialization}
The formation set is initialized by a combination of multiple random formations (rnd) and predefined shaped formations ($\Delta$). This initial set of formations is termed as \textit{'Initialization Pool'}, or $\mathcal{F}^0$, 
\begin{equation}
\mathcal{F}^0 = \mathcal{F}_{\text{rnd}} \cup \mathcal{F}_{\Delta}.
\end{equation}
In the random formations, the position of all the robots is initialized 
using a uniform distribution within the boundaries of the image space. 
Besides, to facilitate a 'warm start' for the optimization algorithm, five predefined basic shapes (rhombus, triangle, inverted triangle, hexagon, square), along with $p$ variations of these shapes (by adding some noise to the positions of the robots) are added to the set.
The parameter $\alpha$ is constant and equal for all the formations during the initialization.
Figure \ref{fig:predefined} shows some examples of these predefined formations along with some random shapes (column on the right) for clarification.

\paragraph{Update}
The formation set is refined at each iteration based on an 'exploration-exploitation' technique, discarding formations with low CLIP similarity and introducing new formations based on random locations (exploration) and different variations of those with good scores in the past (exploitation). In particular,
\begin{equation}
    \mathcal{F}^{k+1}=\mathcal{F}^k_{\text{best}} \cup \mathcal{\tilde{F}}^k_{\text{best}} \cup \mathcal{F}_{\text{rnd}}^{k+1},
\end{equation}
where $\mathcal{F}^k_{\text{best}}$ are the best $b$ formations at iteration $k$, according to the CLIP similarity of their images, $\mathcal{F}_{\text{rnd}}^{k+1}$ is a new subset of random formations, created analogously to the initialization step, and $\mathcal{\tilde{F}}^k_{\text{best}}$ are variations over the best set that we describe next.

We implement four different variations over  $\mathcal{F}^k_{\text{best}}$, incorporating a mix of aggressive and smooth modifications. The resulting formations are incorporated into the new set of formations for the next iterations,
\begin{equation}
\mathcal{\tilde{F}}^k_{\text{best}} =
\mathcal{F}_{\text{subd}} \cup  \mathcal{F}_{\text{one}}  \cup  \mathcal{F}_{\text{contour}} \cup \mathcal{F}_{\alpha}. 
\end{equation}
The first set (subd) involves dividing the map into four halves (top, bottom, left, right). The positions of the robots within each subdivision are altered randomly, following a uniform distribution within small boundaries. 
The second set (one) considers smooth alterations, generated by moving a single robot for all the best formations. 
The third modification (contour) of the formations entails relocating equally all robots to the contour of the shape, introducing a slight amount of noise for further variation.
Finally, in $\mathcal{F}_{\alpha},$ for each formation in $\mathcal{F}^k_{\text{best}}$, we include a new element where we replace the default value of $\alpha$ by the limit value $\alpha_f$ of that formation, leading to the creation of distinct contour drawings in the next iteration.
Once the set is complete, a new iteration begins. The process is repeated for a fixed number of iterations, finally returning the best formation, Eq.~\eqref{Eq:OptimalForm}, along with the values of $\alpha$ and $c^*$.

\subsection{From shapes to drone shows}
\label{sec_postprocessing}

As final step, we need to create the final 3D positions for the drones, decide which drones goes where and move them safely avoiding collisions.

\begin{figure}[!h]
\centering
\begin{tabular}{cc}  
\includegraphics[width=0.30\columnwidth]{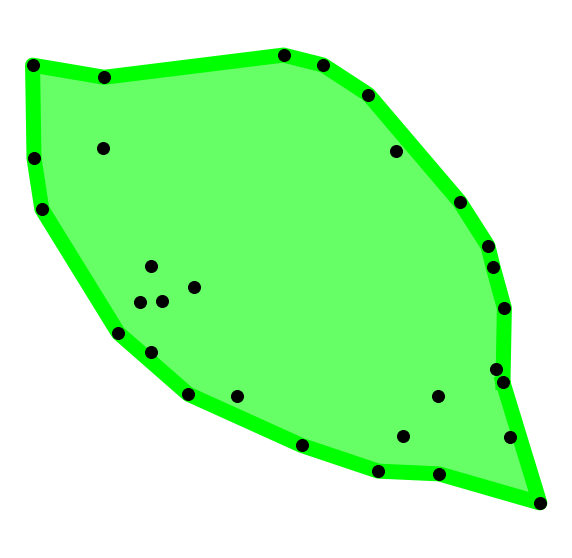}&
    \includegraphics[width=0.32\columnwidth]{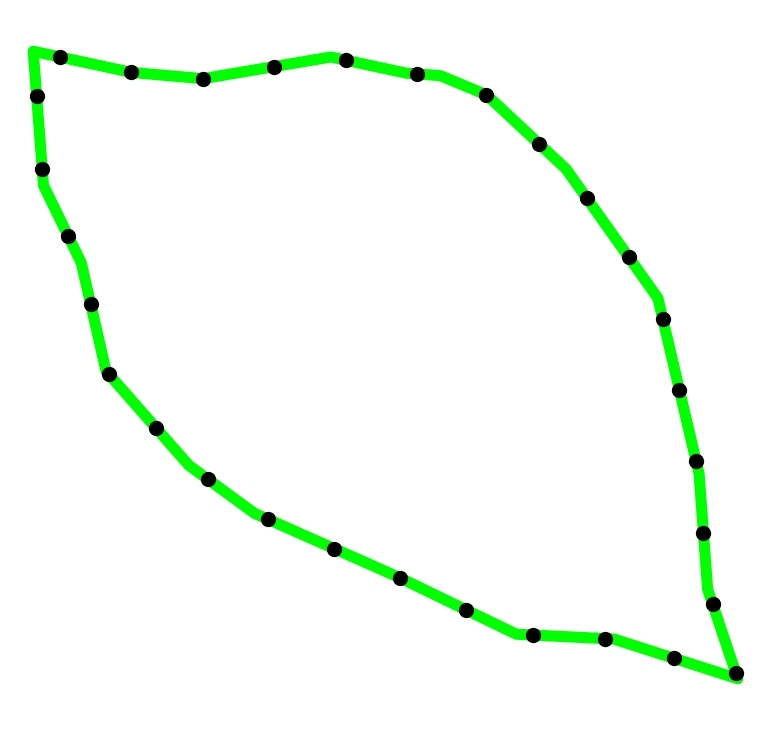}
\end{tabular}
\caption{\footnotesize{\textbf{Postprocessing step to determine the position of the robots}.} The postprocessing step determines the positions of the robots to meaningfully represent the given shape. On the left are the end positions of the 30 robots calculated by the second step of the algorithm. On the right, the postprocess step of the algorithm distributes equally the same number of robots to better represent the same shape. }
\label{fig:alpha_shapes}
\end{figure}

At this stage, we focus only on the contour of the best formation, $f^*$, obtained using the associated value of alpha, $\alpha^*$.
The contour is divided into $M$ equal and joint segments, placing one robot at the extremes of them (Figure~\ref{fig:alpha_shapes}).
Since the formation is in 2D and the drones need to move in 3D, we reproject the formation to be at a fixed distance and height with respecto the the point of view of the spectators.

We consider that initially the drones are equally spaced in the ground. To decide the position in the formation that each drones needs to reach, we use the Hungarian algorithm, minimizing the total distance traveled by the swarm to reach the target formation and obtaining an assignment with low probability of collision.
Finally, to control the drones we use the well-known Optimal Reciprocal Collision Avoidance (ORCA) algorithm for 3D vehicles~\cite{snape2010navigating}.
We note that this process can be repeated with additional formations without the need to start from the ground if needed.

%% file: 04_Experiments.tex
This section demonstrates CLIPSwarm's ability to generate formations that match a provided natural language word, highlighting the applicability of the system to a robotic environment.
A first experiment analyzes the proper behavior of the approach, demonstrating how the algorithm improves CLIP similarity across iterations.
Then, we present a set of experiments to run the whole formation generation pipeline given our defined \textit{test-set} of words. Finally, we demonstrate several executions
of the system in a realistic drone show simulation using AirSim, to showcase the system's applicability to a realistic robotic environment.
The supplementary video provides a more detailed execution of this simulation.

\subsection{Assessing the algorithm}

This experiment investigates how our algorithm progressively increases the CLIP similarity obtained across iterations. 
ensuring the resulting robot formation accurately reflects the target word.

To this end, we define a \textit{test-set} of 50 words \footnote{ \footnotesize {apple, avocado, balloon, banana, bear, bicycle, bird, boat, book, bottle, butterfly, car, cat, chair, cherry, cloud, coin, cup, diamond, drop, fish, flower, hat, heart, house, key, kite, leaf, lemon, lighthouse, lightning, moon, mushroom, orange, pear, plane, puzzle, raindrop, robot, rocket, shoe, spoon, star, strawberry, sun, tomato, train, tree, wave, watermelon}}
representing various shapes.
Subsequently, we execute the first two modules
of the solution (Sec. \ref{sec_algorithm}) for the \textit{test-set}. After conducting several tests, we noticed that the CLIP Similarity metric stops improving after a certain number of iterations. Therefore, for this experiment, we opted for an 'early-stopping' strategy and concluded the training at 15 iterations. 
Formations were generated using $M=30$ robots, and parameters were configured to produce $N=500$ formations in each iteration. 

For all tests, we analyze how the CLIP Similarity keeps growing as our algorithm runs more iterations. In particular, we compute the average percentage of improvement $AoI$ as
\begin{equation}
    P_w = \frac{max(\mathcal{S}^{it_{max}}_w) - max(\mathcal{S}_w^{1})}{max(\mathcal{S}_w^{1})} ,
\end{equation}
\begin{equation}
    AoI = avg(P_w) \quad \forall \quad w \in test\_set, 
\end{equation}
\noindent 
where $S_w^{k}$ represents the list of CLIP Similarities obtained for all formations corresponding to a particular word $w$ at iteration $k$. $P_{w}$ denotes the percentage of improvement, comparing the best CLIP Similarity achieved in the last iteration with that of the first iteration for the word $w$.
\begin{figure}[!h]
\centering
\begin{tabular}{c}  
    \includegraphics[width=0.9\columnwidth]{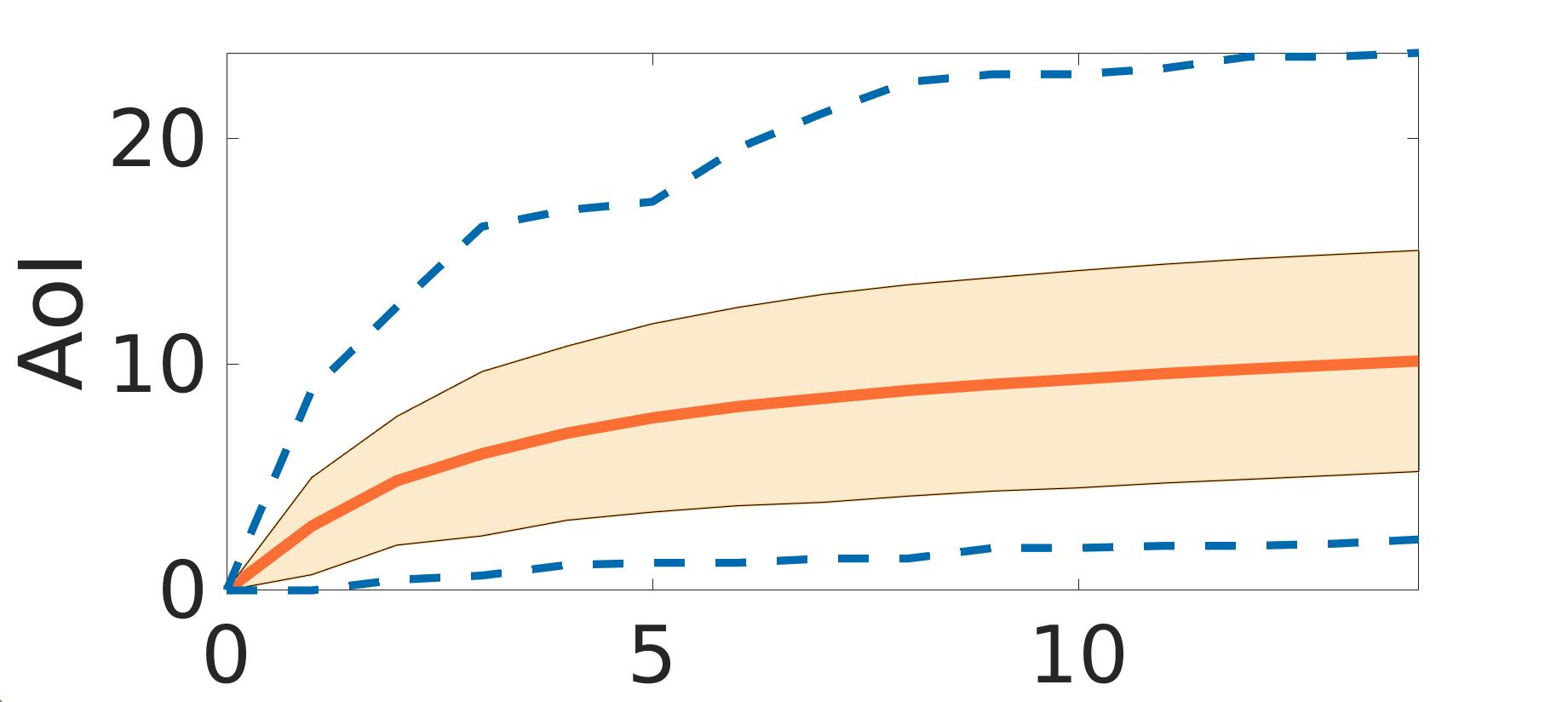}
\end{tabular}
\caption{\textbf{Average of improvement of CLIP Similarity} Solid orange line represents the average improvement of CLIP across iterations for our test set composed of 50 words. Light-shaded area indicates the standard deviation of the average improvement. Top dashed line represents a particular word where the average improvement increases the most across iterations. The bottom dashed line represents a particular word where the average improvement increases the least across iterations, possibly due to a good seed.}
\label{fig:graph}
\end{figure}

A graphical representation of the results of this test after executing the solution for the 50 words of the \textit{test-set} is depicted in Fig.\ref{fig:graph}. The solid line illustrates the average percentage of improvement across iterations relative to the initial maximum CLIP similarity. At the end, the $AoI$ for all words is 10.15\% with respect to the first iteration. The lighter-shaded area represents the standard deviation. The dashed lines at the top and bottom indicate the scenarios where CLIP similarity demonstrates the most improvement across iterations and the worst improvement (meaning that the initial shape is a good fit for the word), respectively.

\begin{figure}[!t]
\centering

{\small \textbf{User input 1: ``Cat"}}\\
\vspace{-0.3em}
{\footnotesize \textbf{Enriched input 1: ``A magenta cat shape"}}

\begin{tabular}{cccc}
   \includegraphics[width=0.2\linewidth]{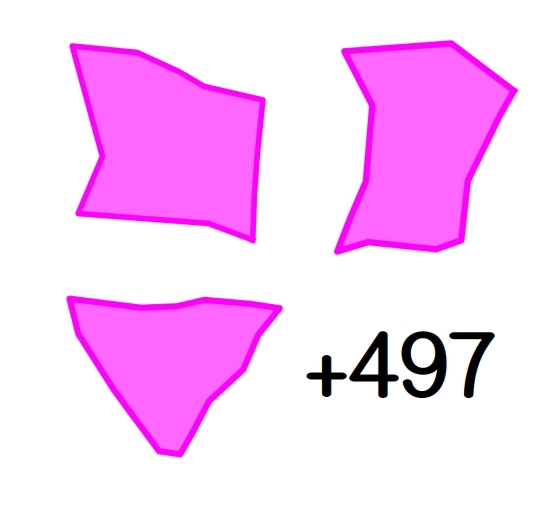}
    & \includegraphics[width=0.2\linewidth]{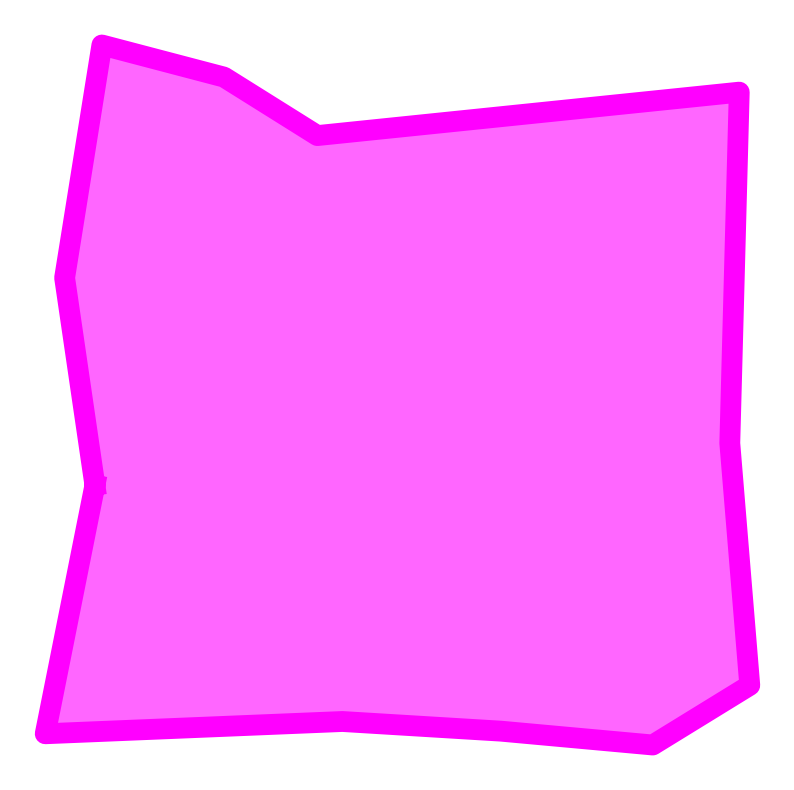}
    & \includegraphics[width=0.2\linewidth]{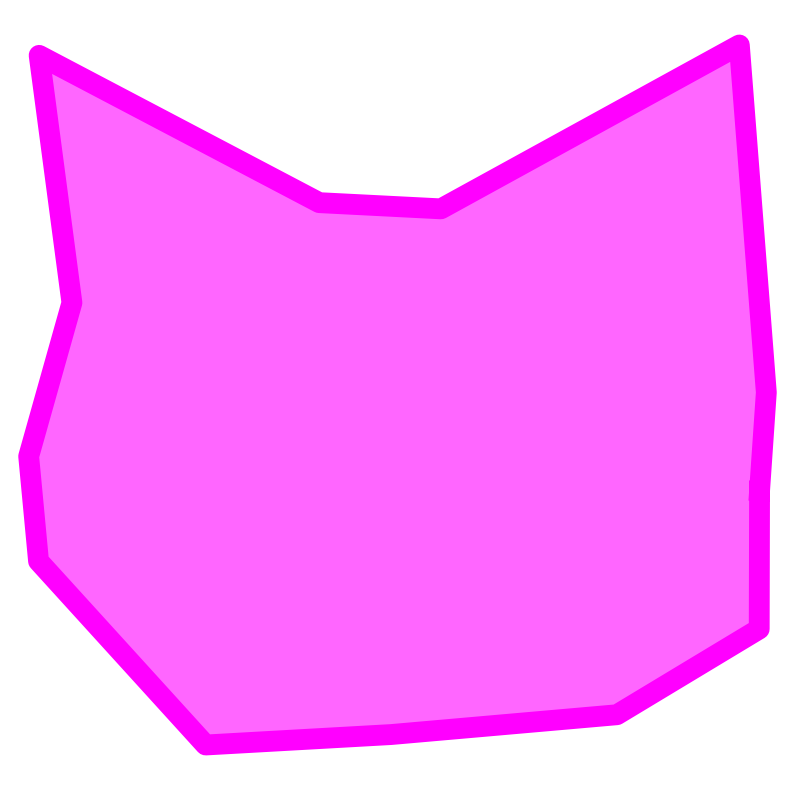}
    & \includegraphics[width=0.2\linewidth]{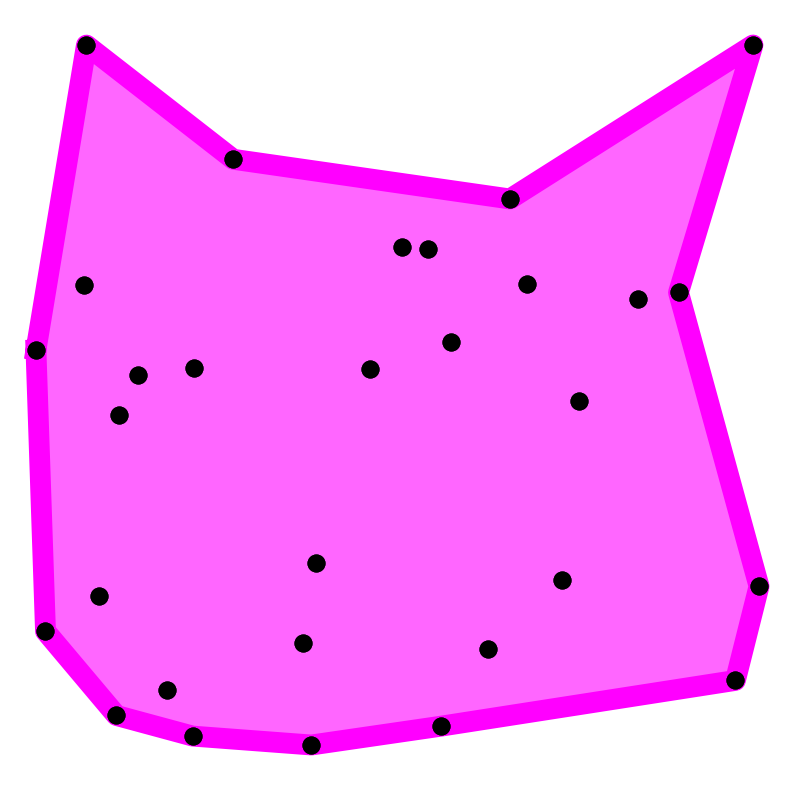}
    \\
    
\end{tabular}

{\small \textbf{User input 2: ``Lemon"}}\\
\vspace{-0.3em}
{\footnotesize\textbf{Enriched input 2: ``A yellow lemon shape"}}

\begin{tabular}{cccc}
   \includegraphics[width=0.2\linewidth]{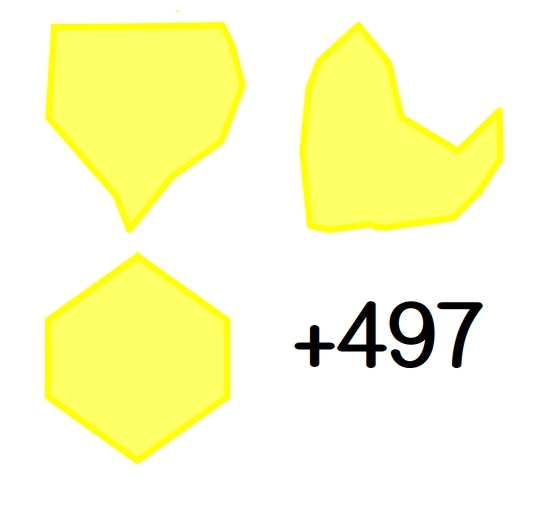}
    & \includegraphics[width=0.2\linewidth]{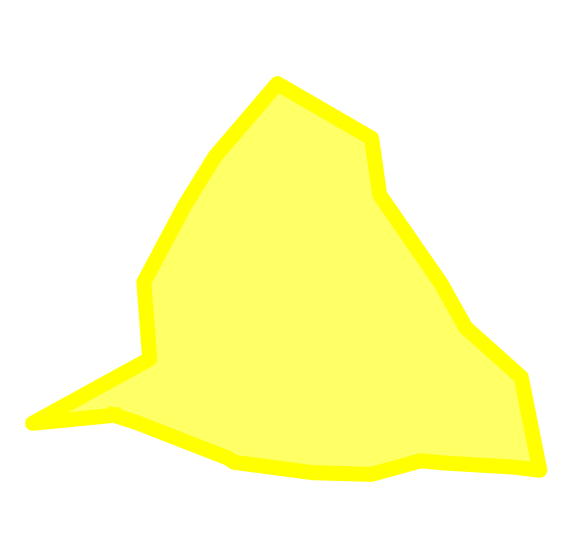}
    & \includegraphics[width=0.2\linewidth]{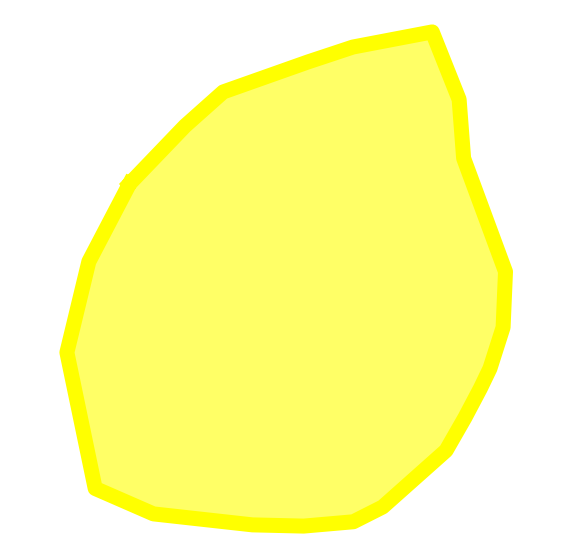}
    & \includegraphics[width=0.2\linewidth]{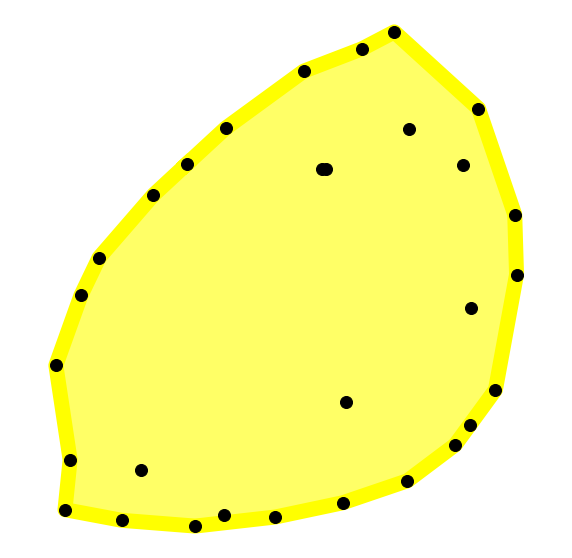}
    \\ 
\end{tabular}
{\small\textbf{User input 3: ``Apple"}}\\
\vspace{-0.3em}
{\footnotesize\textbf{Enriched input 3: ``A red apple shape"}}

\begin{tabular}{cccc}
   \includegraphics[width=0.20\linewidth]{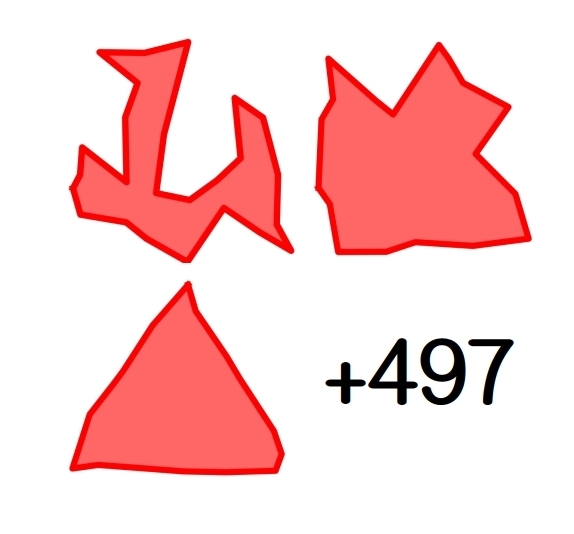}
    & \includegraphics[width=0.20\linewidth]{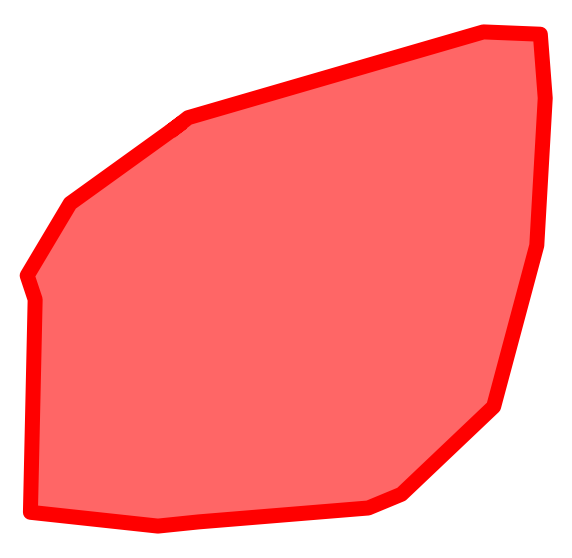}
    & \includegraphics[width=0.20\linewidth]{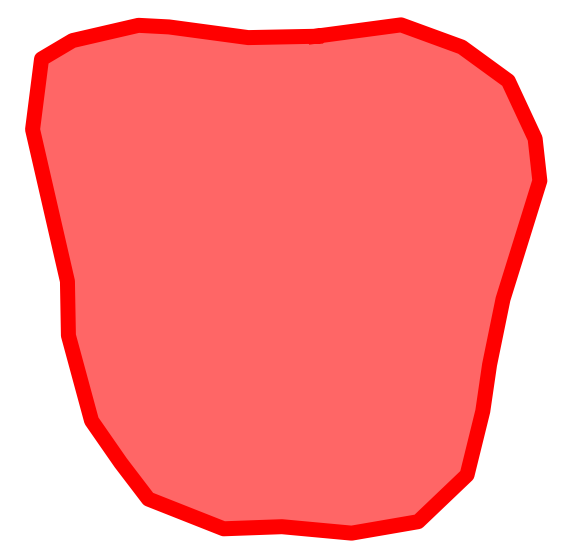}
    & \includegraphics[width=0.20\linewidth]{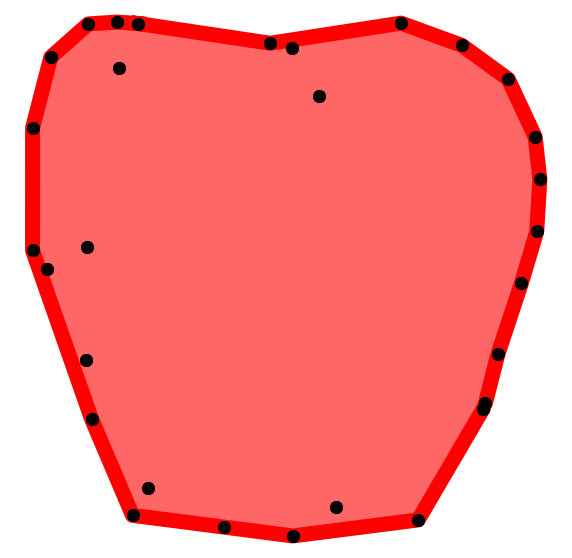}
    \\ 
\end{tabular}

{\small \textbf{User input 4: ``Raindrop"}}\\
\vspace{-0.3em}
{\footnotesize\textbf{Enriched input 4: ``A cyan raindrop shape"}}

\begin{tabular}{cccc}
   \includegraphics[width=0.2\linewidth]{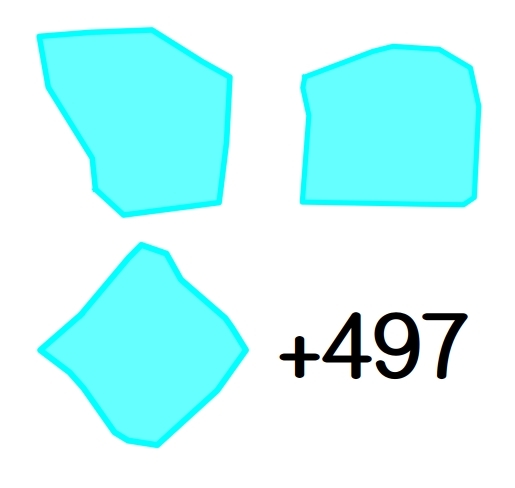}
    & \includegraphics[width=0.2\linewidth]{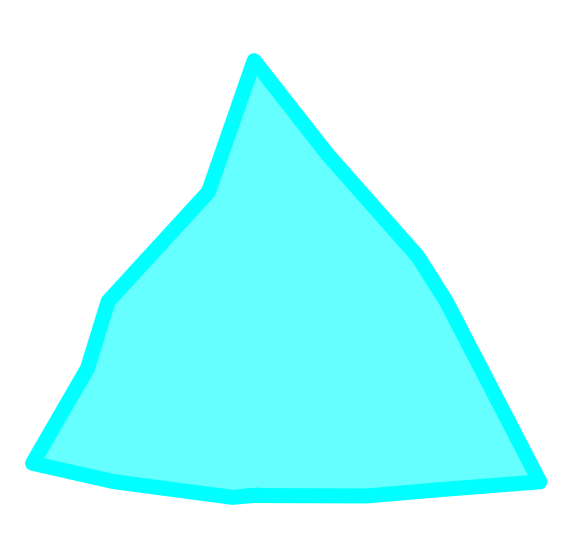}
    & \includegraphics[width=0.2\linewidth]{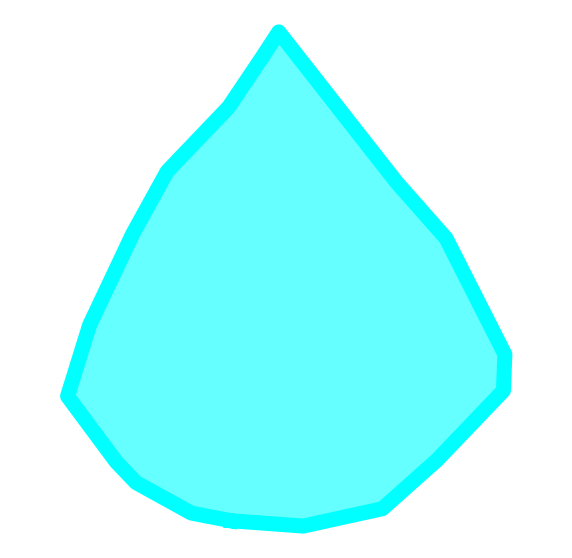}
    & \includegraphics[width=0.2\linewidth]{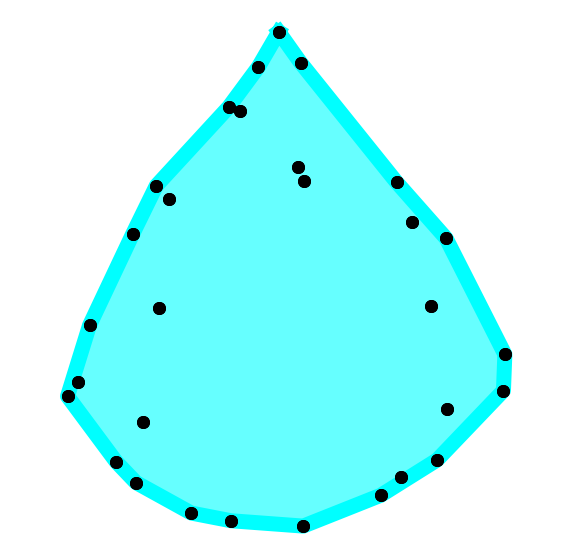}
    \\ 
\end{tabular}

{\small \textbf{User input 5: ``Heart"}}\\
\vspace{-0.3em}
{\footnotesize \textbf{Enriched input 5: ``A magenta heart shape"}}
\begin{tabular}{cccc}
   \includegraphics[width=0.2\linewidth]{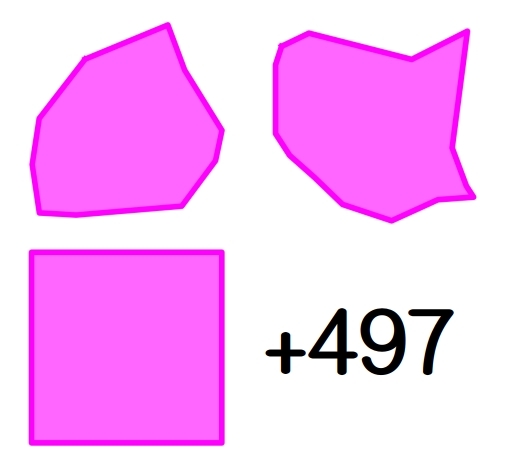}
    & \includegraphics[width=0.2\linewidth]{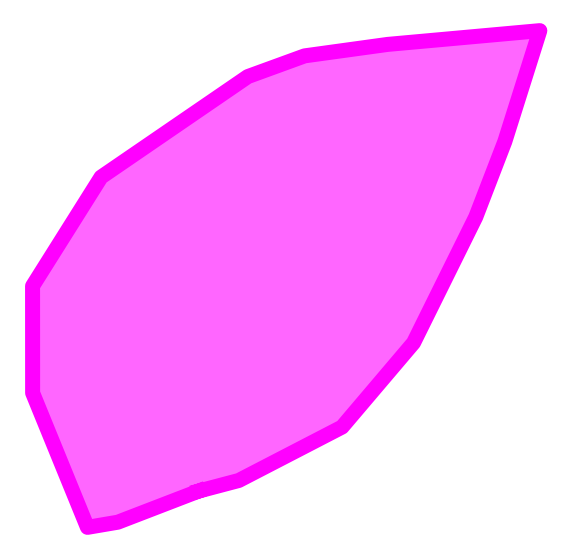}
    & \includegraphics[width=0.2\linewidth]{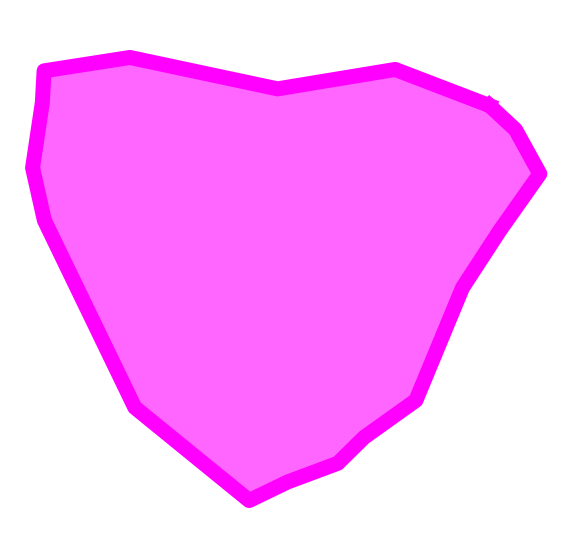}
    & \includegraphics[width=0.2\linewidth]{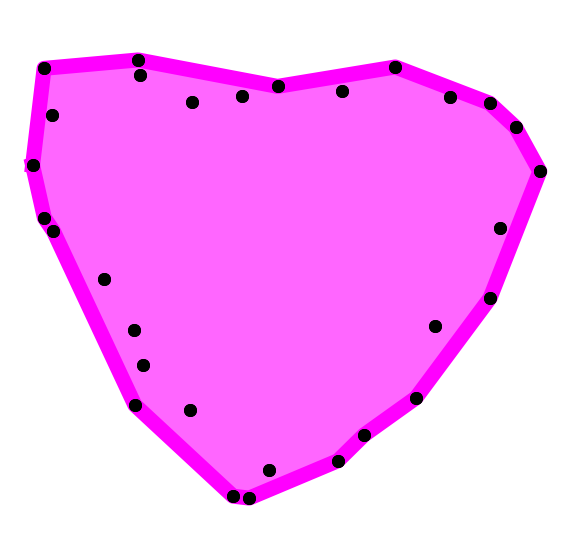}
    \\

\end{tabular}
    {\small \textbf{User input 6: ``Leaf"}}\\
\vspace{-0.3em}
{\footnotesize\textbf{Enriched input 6: ``A green leaf shape"}}

\begin{tabular}{cccc}
   \includegraphics[width=0.2\linewidth]{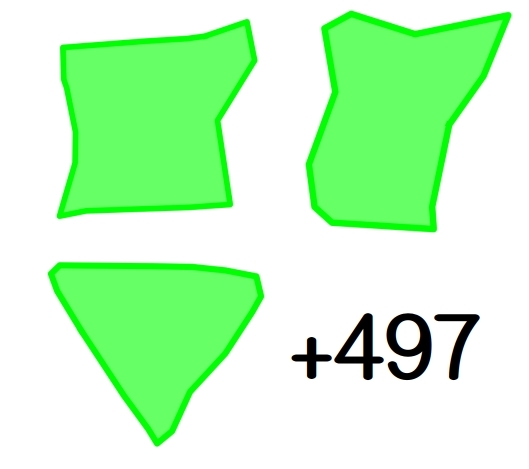}
    & \includegraphics[width=0.2\linewidth]{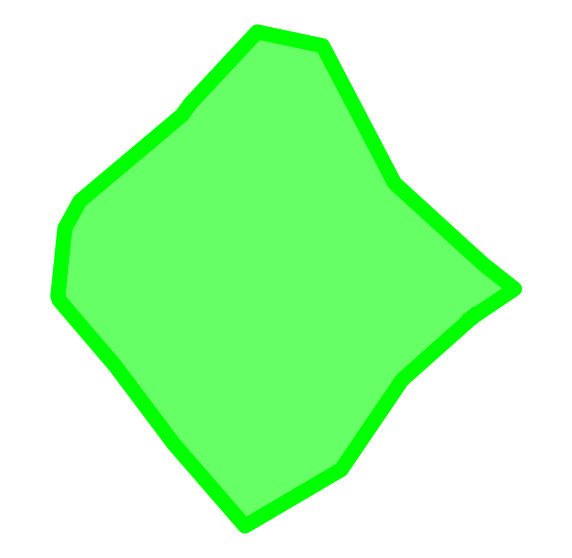}
    & \includegraphics[width=0.2\linewidth]{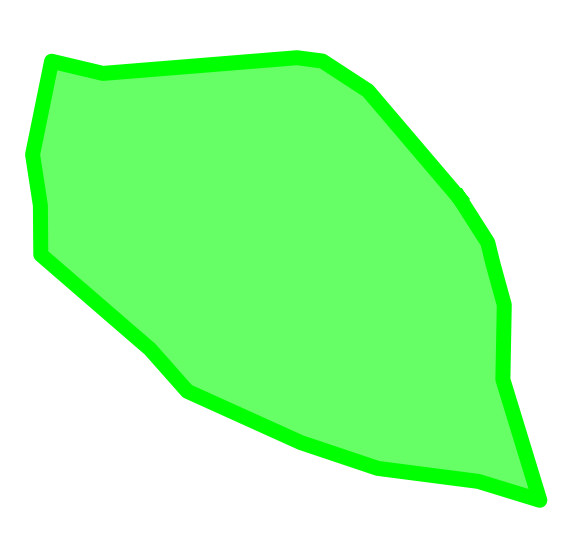}
    & \includegraphics[width=0.2\linewidth]{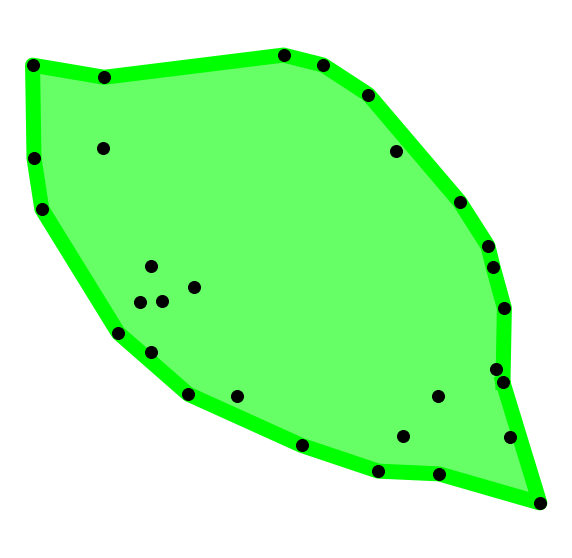}
    \\
    \footnotesize{Initialization Pool} &\footnotesize{it = 1} &\footnotesize{it = 7} &\footnotesize{it = 15} 
    \\ 
    \\ 
\end{tabular}
\caption{\textbf{From a Single Word to Shapes.} 
Shapes crafted from formations of 30 robots. The first column displays the representation of the initialization pool used as the seed in the Initialization step, composed of random and predefined shapes for a 'warm start.' Subsequent columns show shapes with the highest similarity in different iterations (1, 7, 15) of the algorithm, given the input text and enriched description detailed in the title of each row. The fourth column represents the formation with the highest similarity after 15 iterations, also chosen to be used in the drone show, since it best matches the input word. In this case, the positions of the robots of the formations are depicted for comparison with their positions in Fig.~\ref{fig:drone_show}, after the equal distribution of robots along the contour, as shown in Fig.~\ref{fig:alpha_shapes}.
}
\label{fig:formations}
\end{figure}

\subsection{Modeling formations from a word}
This section showcases formations generated by the algorithm in response to an input word and their evolution over successive iterations. 
For this experiment, we chose certain shapes from the outcomes of the previous experiment, after running it on the \textit{test-set}. These shapes will be utilized for the drone show in the subsequent experiment.
Figure~\ref{fig:formations} illustrates the results of this experiment. Each row corresponds to one test with one word. It depicts the representation of the formation with the highest score at various iterations of the algorithm, in response to the introduced word marked as 'User input.' The algorithm dynamically enriches the prompt, shown in the figure as 'Enriched input.' 
The initial figure in each row displays a representation of the initialization pool that is used as a seed in the Initialization step and just composed by random shapes.
For quantitative results, Table \ref{table:similarity} shows the best CLIP Similarity for each case for the initial, intermediate, and final iteration. This metric increases as the iterations progress, indicating the solution's ability to obtain formations that match closer and closer the given description.

\begin{table}[!t]
\begin{center}
\caption{Best Similarity Metric/Iteration for Shapes in the Drone Show}
\begin{tabular}{|c|c|c|c|c|c|c|}
\hline
& Cat & Lemon & Apple & Raindrop & Heart  & Leaf \\ \hline
it=1 & 0.250 & 0.305 & 0.245 & 0.265 & 0.270 & 0.310 \\ 
\hline
it=7 & 0.278 & 0.329 & 0.284 & 0.292 & 0.280 & 0.317 \\ 
\hline
it=15 & 0.281 & 0.334 & 0.297 & 0.296 & 0.302 & 0.325 \\ 
\hline

\end{tabular}
\label{table:similarity}
\end{center}
\end{table}

\subsection{Performing a Drone Show in photorealistic simulation}
This section emphasizes the adaptability of the system to a realistic robotic environment, highlighting the effect of the postprocessing stage of our algorithm (Sec.\ref{sec_postprocessing}).
To demonstrate this, we conducted a drone show in the photorealistic simulator AirSim \cite{airsim, pueyo2020cinemairsim}, offering a scenario closer to reality. 
To specify the color of the drones, we equipped them with a light whose color can be controlled through the AirSim API. The connection between CLIPSwarm and AirSim is implemented in ROS \cite{ros}, facilitating the potential transition of the solution to a setup with real drones. 

We employed a Python adaptation \cite{orca_lib} of ORCA for 3D formations to send control commands to the drones. This algorithm ensures that the drones progress towards the goals in smaller steps to reach the final position, avoiding collisions between them.

\begin{figure}[!t]
\centering
\begin{tabular}{cc}  
\includegraphics[width=0.46\columnwidth, height=2.5cm]{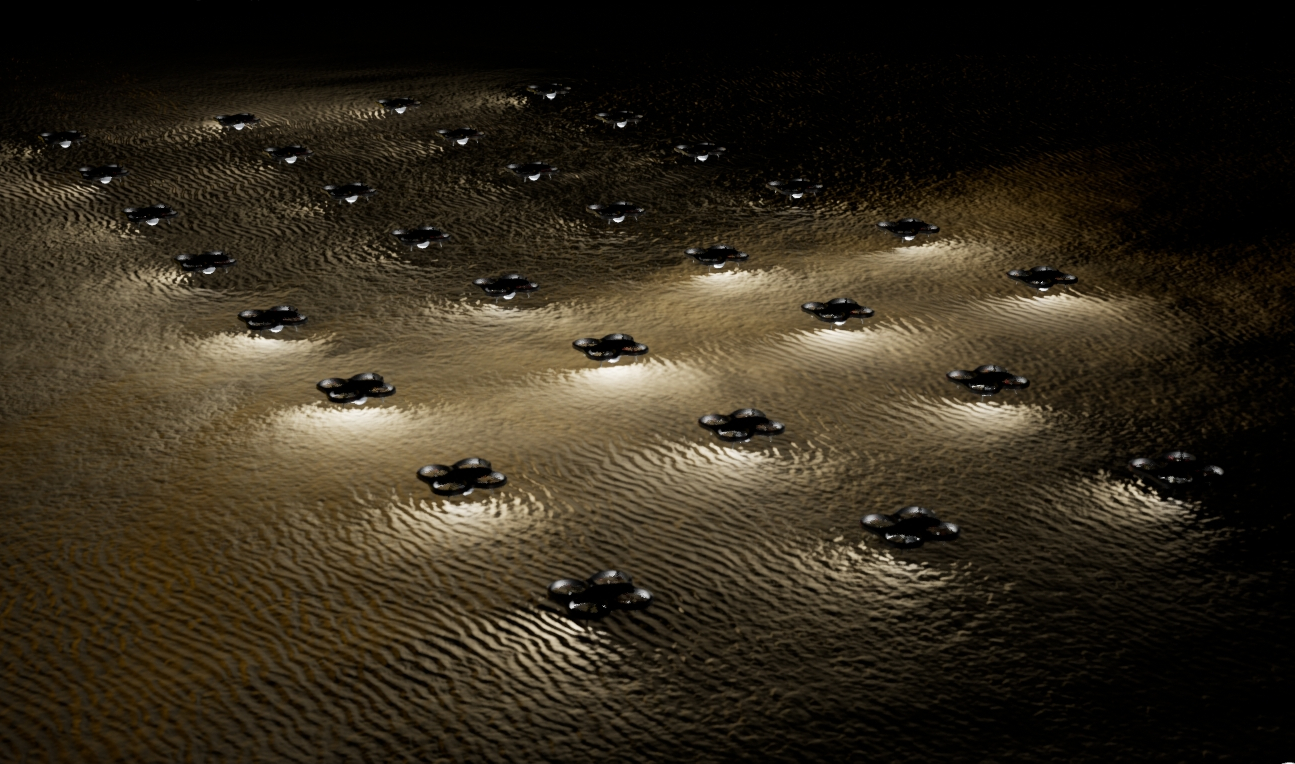}
&    
 \includegraphics[width=0.46\columnwidth, height=2.5cm]{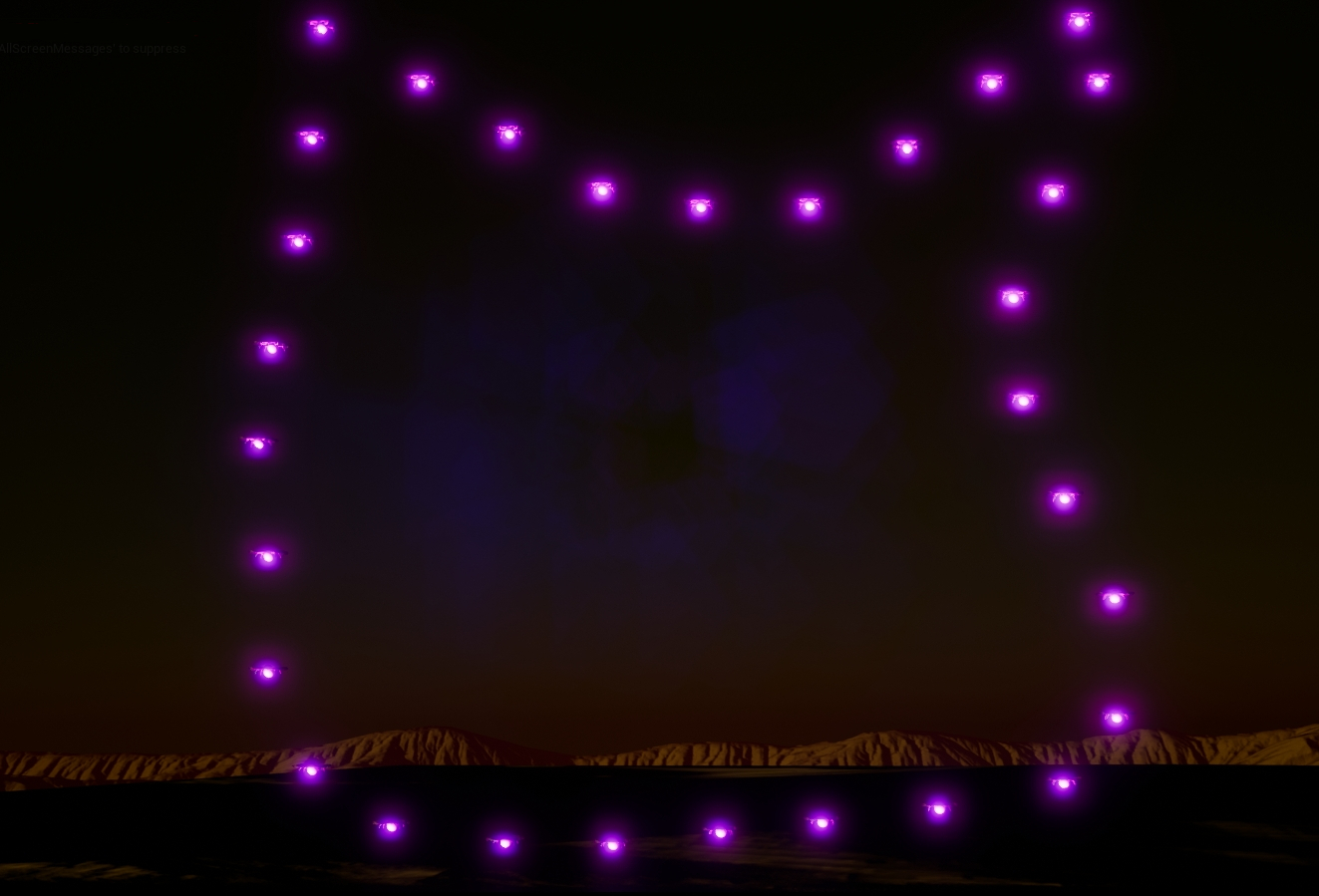}
\\
 \includegraphics[width=0.46\columnwidth, height=2.5cm]{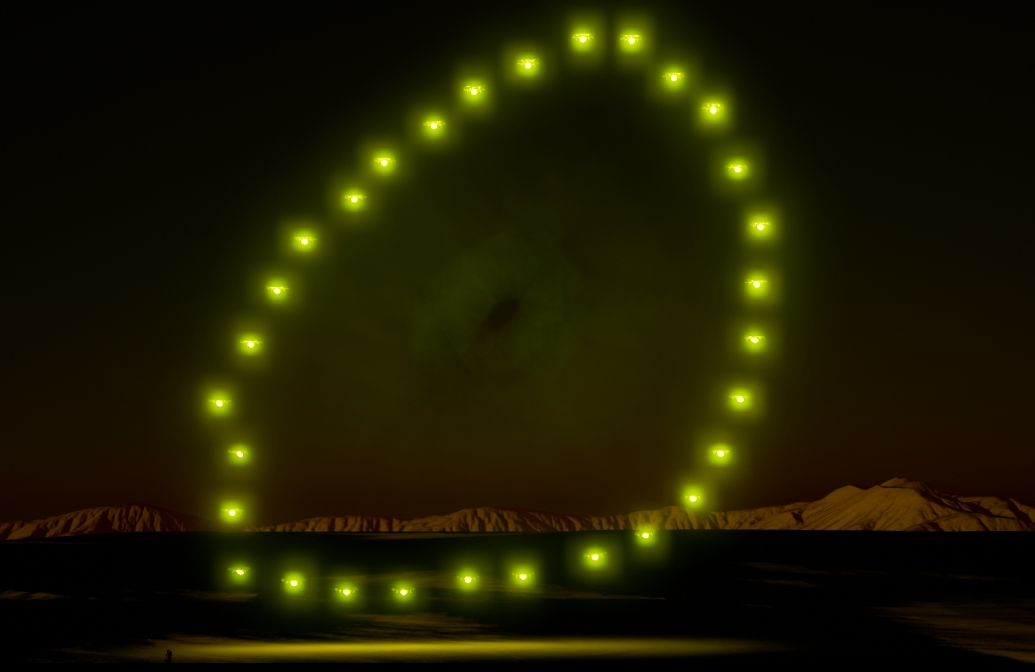}
 &
\includegraphics[width=0.46\linewidth, height = 2.5cm]{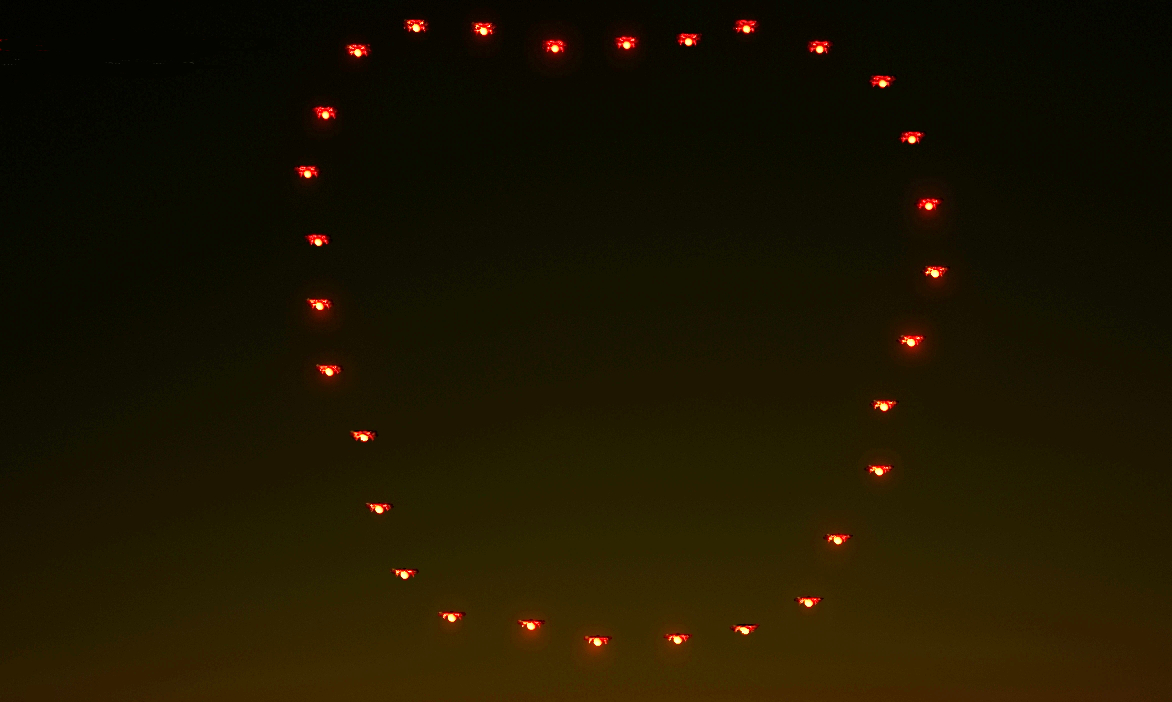}
\\
    \includegraphics[width=0.46\linewidth, height = 2.5cm]{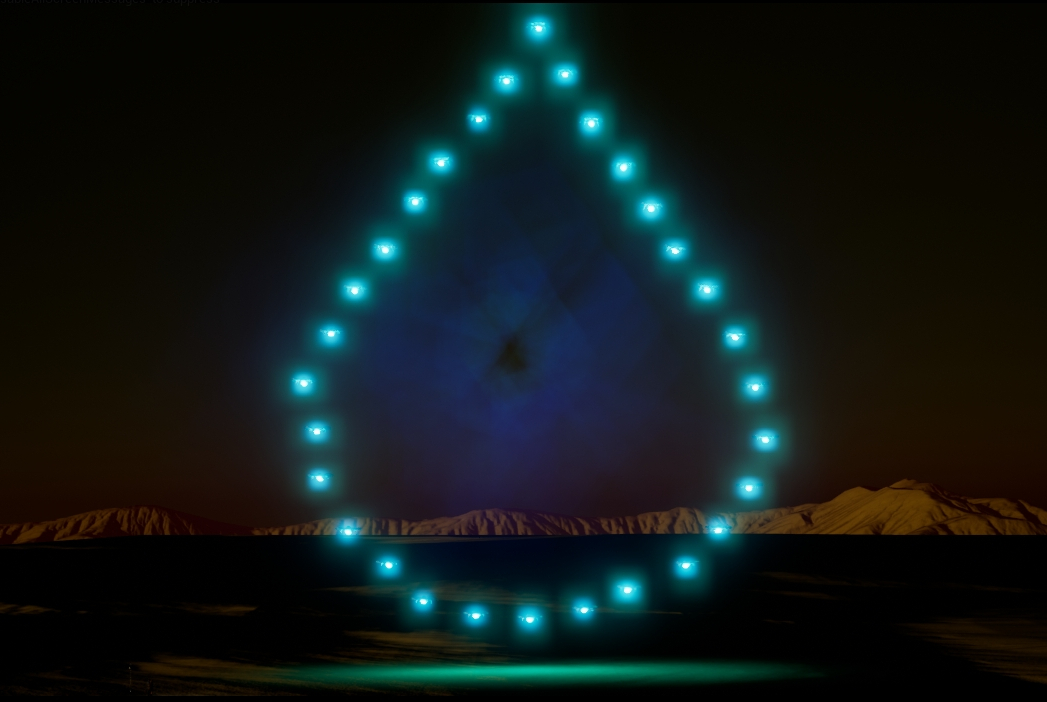}
&
    \includegraphics[width=0.46\linewidth, height = 2.5cm]{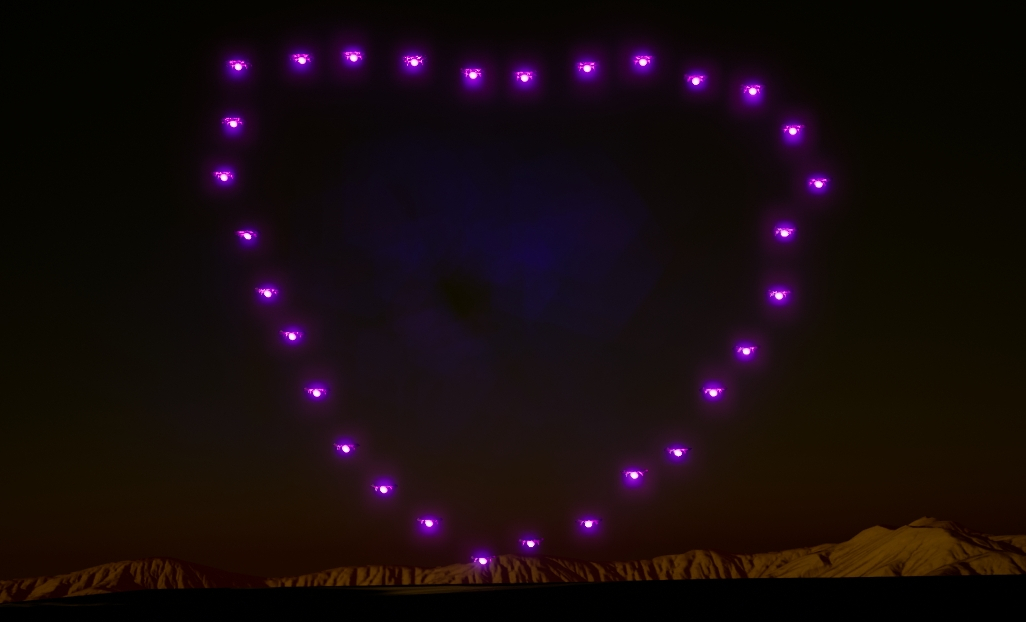}  
\\
    \includegraphics[width=0.46\linewidth, height = 2.5cm]{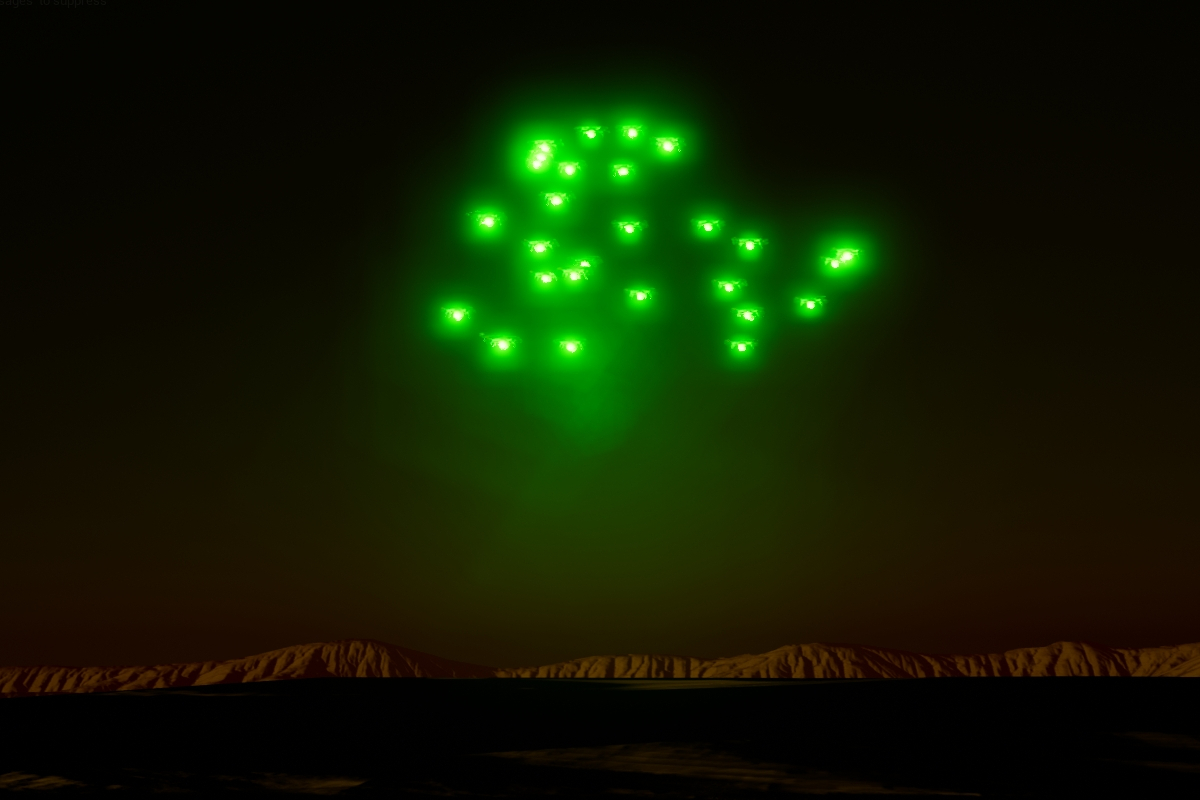}    
    &
    
    \includegraphics[width=0.46\linewidth, height = 2.5cm]{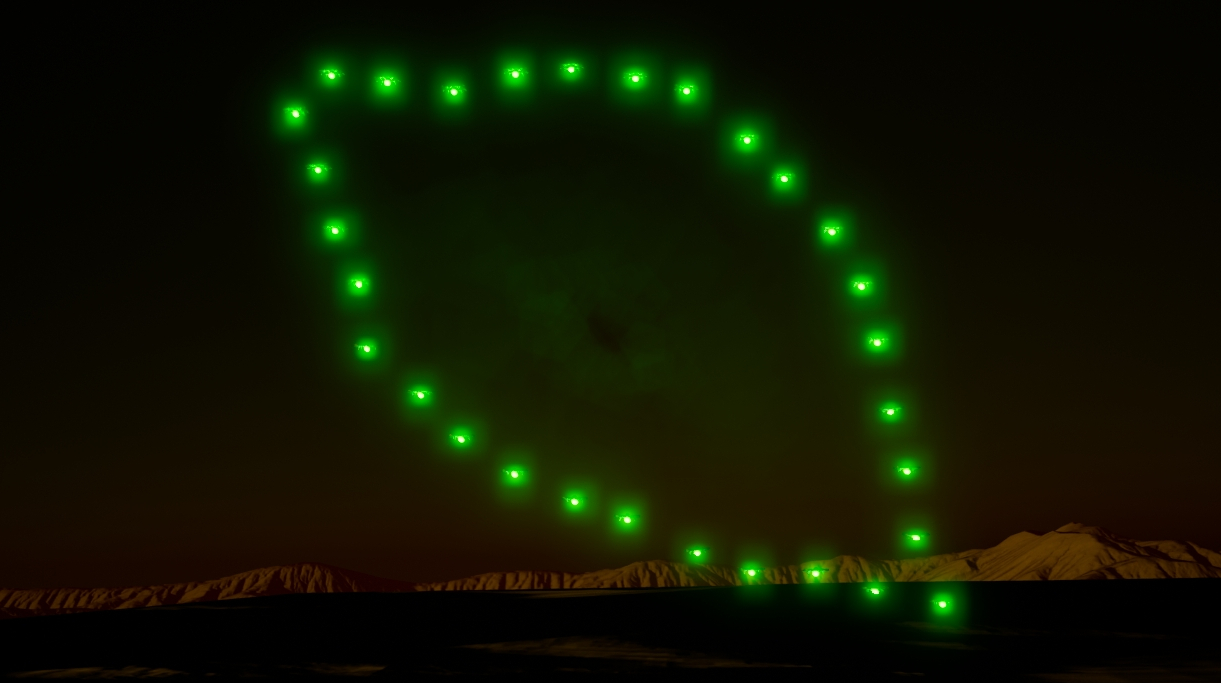}

\end{tabular}

\caption{\textbf {Performing a Drone Show in photorealistic simulation}. Frames from photorealistic simulation showcase different stages of a drone show, arranged chronologically left to right. The shapes have been automatically created in earlier stages of the algorithm when the next subset of words from the \textit{test-set} are inputted: ['cat', 'lemon', 'apple', 'raindrop', 'heart', 'leaf']. The show is performed in a photorealistic simulation (AirSim) with 30 drones. The initial position of the drones is depicted on the first row on the left. An intermediate step is depicted in the bottom row on the left to demonstrate the transition of drones between shapes, and their ability to respect dynamics and avoid collisions between drones. }
\label{fig:drone_show}
\end{figure}

The stages of the drone show in photorealistic simulation are depicted in Fig. \ref{fig:drone_show}. The shapes are automatically obtained using CLIPSwarm. The drone show replicates the shapes obtained in the last iteration of the algorithm, as shown in the fourth column of Fig. \ref{fig:formations}. In comparison with these formations, the postprocessing stage distributes the drones equally along the contour and reprojects the shapes from 2D to 3D, as described in Section \ref{sec_postprocessing}.
The figure displays the initial positions of the drones (top left). An intermediate stage of the drone show is displayed (bottom left) to represent the transition from one shape to another. The postprocessing algorithm ensures drone dynamics and collision avoidance.
We direct the reader to the supplementary video for a complete visualization of the drone show, showcasing all the transitions and shapes.

%% file: 05_Limitations.tex
\begin{figure}[!b]
\centering
{\small\textbf{User input 1: ``House"}}\\
\begin{tabular}{cc}  
\includegraphics[width=0.25\columnwidth]{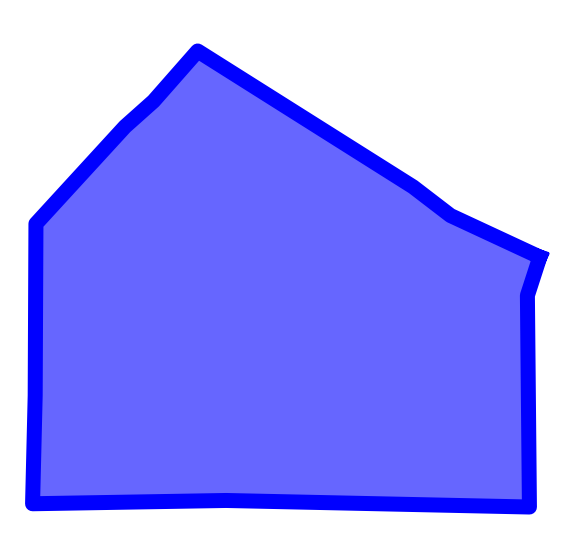}&
    \includegraphics[width=0.25\columnwidth]{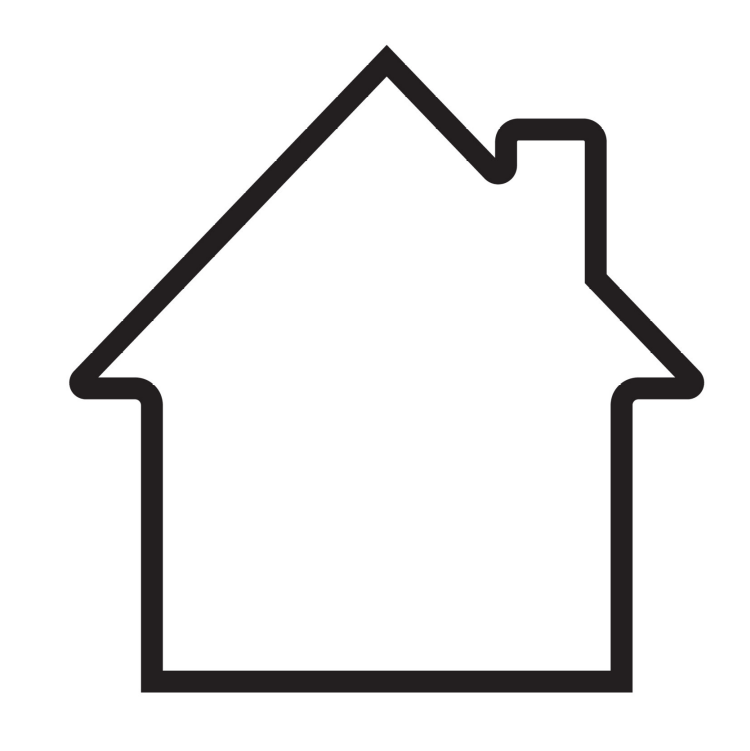}
    \\ Clip Similarity =  0.3062 & Clip Similarity = 0.2881
\end{tabular}
\caption{\footnotesize{\textbf{Limitations of the algorithm}.} This example illustrates the limitations of the algorithm. On the left, the shape obtained by our algorithm as a response to the input text "House". On the right, a stock example of a contour of a house. The contour simplifies the evaluation process but may not capture all the details of the formation's shape. Additionally, relying solely on the CLIP Similarity metric may result in a lower score for formations that better correspond to the given text for an average user. }
\label{fig:lim}
\end{figure}

For the sake of simplicity, we have decided to evaluate the formations based on the CLIP Similarity of the Alpha-shape contour of the formations. This simplification allows for a faster evaluation process and works well with a relatively low number of robots. However, the variety of shapes that can be modeled with just a contour is limited. Moreover, our current algorithm relies significantly on CLIP, and increasing the similarity does not always mean that the shape is closer to what an average user would expect, as CLIP is not specifically trained with contours.
We illustrate this with an example in Fig.~\ref{fig:lim}, where the algorithm fails to capture all the expected details that would represent a house. In this case, the CLIP Similarity of the picture on the left is higher than the one on the right, even though the latter may seem closer to a \textit{house} for an average reader, showcasing a limitation of using this CLIP as a similarity metric. Future steps will include working with more complex inputs as well as using additional metrics that complement the CLIP Similarity for a more insightful comparison of the images and texts.

%% file: 06_Conclusions.tex
In this paper, we introduced CLIPSwarm, which is designed to create automatically drone formations that represent a given word in natural language. We have explained how we enriched the word provided and engineered a corresponding text prompt. This text is then used by an iterative algorithm that employs an 'exploration-exploitation' technique to derive a formation of robots that aligns with the description in the given text. We used CLIP to encode the text and the images into vectors to measure the similarity between the description and the image of the formations. The formation is then adapted to visually represent the word within the constraints of the available number of drones. Control actions are assigned to the drones to guide them to their positions, ensuring robotic behavior and collision-free movement.

Our experiments have shown that the algorithm improves the clip similarity across iterations. Additionally, the system can model robot formations from natural language and be applied to performing drone shows featuring different shapes and colors. In the future, we will explore a wider variety of formations, including more complete shapes and the optimization of 3D formations.